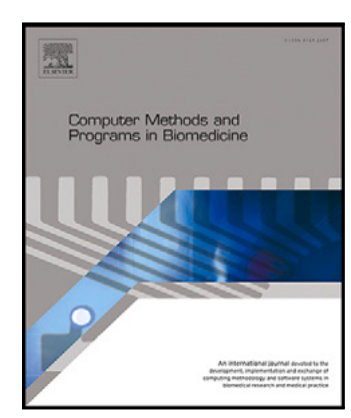

# Physiological-model-based neural network for modeling the metabolic–heart rate relationship during physical activities

Yaowen Zhang [a], Libera Fresiello [b], Peter H. Veltink [a], Dirk W. Donker [b,c], Ying Wang [a,*]

[a] Department of Biomedical Signals and Systems, University of Twente, Enschede, 7522NB, The Netherlands
[b] Department of Cardiovascular and Respiratory Physiology, University of Twente, Enschede, 7522NB, The Netherlands
[c] Department of Intensive Care, University Medical Center Utrecht, Utrecht, 3584CX, The Netherlands

## ARTICLE INFO

## ABSTRACT

**Background and Objective:** Heart failure (HF) poses a significant global health challenge, with early detection offering opportunities for improved outcomes. Abnormalities in heart rate (HR), particularly during daily activities, may serve as early indicators of HF risk. However, existing HR monitoring tools for HF detection are limited by their reliability on population-based averages. The estimation of individualized HR serves as a dynamic digital twin, enabling precise tracking of cardiac health biomarkers.

**Methods:** This study introduces a novel physiological-model-based neural network (PMB-NN) framework to model the oxygen uptake ($\dot{V}O_2$)-HR relationship during physical activities, establishing a physiology-grounded intermediate module for future daily life HR estimation from body movement signals. PMB-NN embeds physiological constraints, derived from our proposed simplified metabolic–HR physiological model (PM), into the neural network training process. The framework was trained and tested on individual datasets from 25 participants engaged in activities including resting, cycling, and running. HR estimation performance was evaluated for PMB-NN, comparing with benchmark fully connected neural network (FCNN) and PM, across three dimensions: numerical accuracy, physiological plausibility, and physiological interpretability. Furthermore, sensitivity analysis was conducted to verify the model's robustness against input uncertainty.

**Results:** The PMB-NN model adheres to human physiological principles while achieving high estimation accuracy, with median $R^2$ score of 0.88, RMSE of 9.96 bpm and MAE of 8.87 bpm, even in the presence of intermittent data. Comparative statistical analysis demonstrates that the PMB-NN achieves performance on par with FCNN while significantly outperforming PM ($p < 0.001$). Meanwhile, PMB-NN reaches higher plausibility for HR-$\dot{V}O_2$ coupling ($\rho = 1$) than both FCNN (p = 0.028) and PM ($p < 0.001$). Furthermore, PMB-NN is adept at identifying personalized parameters of the PM, enabling reasonable HR reconstruction. Sensitivity analysis reveals that PMB-NN yields an RMSE within 15 bpm despite input uncertainties of up to 20% Gaussian noise, 4% outliers, and an 18 s time lag.

**Conclusion:** This study confirms the validity of the PMB-NN framework using precise metabolic inputs. This foundational validation enables future integration with wearable-based $\dot{V}O_2$ estimation systems, ultimately paving the way for personalized, real-time cardiac monitoring during daily life physical activities to enhance HF risk detection.

## 1. Introduction

Heart failure (HF) contributes to an escalating global health problem, affecting over 15 million people in Europe and imposing a significant socio-economic burdens on both patients and healthcare systems [1,2]. Early detection of HF would offer opportunities for implementing lifestyle changes and early pharmacological treatments with the great potential to slow the progression of the disease and enhance patient outcomes [3]. Alterations in HR patterns can enable healthcare practitioners to detect warning signs of HF through daily monitoring and promptly conduct further investigations for diagnosis. Elevated resting heart rate (HR) and impaired HR recovery after exercise are strong clinical indicators of HF [4–6]. Additionally, an increasing rate in HR at the onset of exercise is relative to higher cardiovascular mortality, while suppressed peak HR during exercise has been shown to carry significant prognostic value for HF [7]. Given these findings, abnormalities in HR fluctuations are expected to manifest during daily activities in individuals at risk for HF or those with chronic HF, further underscoring the abnormalities' relevance in continuous monitoring and early detection [8–10].

* Corresponding author.
*E-mail addresses:* y.zhang-12@utwente.nl (Y. Zhang), l.fresiello@utwente.nl (L. Fresiello), p.h.veltink@utwente.nl (P.H. Veltink), d.w.donker@utwente.nl (D.W. Donker), imwywk@gmail.com (Y. Wang).

However, existing HR monitoring systems provide limited utility in daily practice as they primarily assess cardiac health by comparing detected HR values with population-based averages [11,12]. These averages generally do not represent individual's normal HR fluctuations especially during different physical activities in daily life. It is therefore essential to establish a real-time personalized HR estimation model that can serve as an individual's digital twin. This digital twin does not replace real-time sensors; rather, it functions as a reference generator, providing an expected HR response baseline based on physiological dynamics. By continuously comparing the sensor-measured actual HR against this digital twin's estimation, healthcare practitioners can distinguish between normal physiological variability and systematic anomalies, enabling more precise tracking of cardiac health status.

Conventionally, HR estimation models consist of a mathematical representation of the cardiovascular circulation and control mechanism response to aerobic exercise with physiology knowledge embedded [13,14]. These models have been significantly refined to replicate the main circulatory subsystem representations within the body, thereby enabling accurate simulations of cardiovascular responses to aerobic exercise. However, sole mathematical models are designed to tailor the HR estimation to specific postures or activity scenarios, for instance, participants were instructed to keep the supine position [13] or cycling at a constant rate of 60 rpm [15]. Beyond the contextual limitations, these models include numerous cardiovascular parameters whose values are typically tuned according to standard population representative values, while their computational efficiency is limited by the complexity arising from the multiple interacting variables. Accordingly, it is challenging and unfeasible to use these models for timely and personalized HR estimation during daily activities.

Data-driven approaches have achieved strong performance in HR estimation, particularly with the increasing availability of wearable data collected in daily-life settings. Existing studies encompass two major architectures: (i) feature-engineering-based machine learning pipelines that rely on manually extracted signal characteristics [16–18], and (ii) end-to-end deep learning pipelines that learn directly from raw physiological signals [19–22]. These methods have been applied across diverse scenarios, from sleep and sedentary postures to daily-life activities and intensive exercise. Our previous work [23] also demonstrated that HR can be reliably estimated from body movement signals using a purely data-driven framework that leveraged temporal HR dependencies to recover HR dynamics under free-living conditions. Although such models achieve high numerical accuracy, their robustness degrades under complex physiological variability, such as during vigorous exercise. Moreover, while temporal HR sequences and HR-dependent signals (e.g., beat-to-beat photoplethysmography or chest-mounted inertial measurement units for heart beats) can enhance estimation performance, they primarily capture correlative temporal patterns rather than the causal mechanisms governing cardiovascular regulation. They describe how HR evolves but not why, as the true physiological drivers — such as metabolic demand, which is the focus of this study — remain unmodeled in purely data-driven approaches.

To overcome the limitations of existing approaches, we hypothesized that a hybrid architecture — integrating simplified physiological principles into a data-driven framework — could leverage the interpretability of mechanistic models while retaining the high estimation accuracy of deep learning. As a critical intermediate step towards enabling daily-life monitoring via wearable sensors, this study focuses on establishing and validating such a framework using precise metabolic measurements. Accordingly, we proposed a simplified physiological model (PM) for estimating HR based on $\dot{V}O_2$ and, building upon this, introduced a hybrid framework: the Physiological-Model-Based Neural Network (PMB-NN). Inspired by emerging Physics-Informed Neural Networks (PINNs) [24–26], the PMB-NN utilizes prior physiological knowledge to constrain data-driven modeling. HR estimation performance was evaluated across three dimensions: numerical accuracy, physiological plausibility, and physiological interpretability. We further specify our hypotheses as follows:

1. the proposed simplified PM can demonstrate a reasonable capability for estimating HR across various activities;
2. the proposed PMB-NN model can elevate the performance of the sole PM;
3. the proposed PMB-NN model can achieve equivalent performance as the sole data-driven model and meanwhile improve its physiological plausibility;
4. the proposed PMB-NN model is capable of identifying physiologically relevant parameters, which are specifically defined and assigned within the context of the model.

## 2. Methods

### 2.1. Dataset and data pre-processing

This study was carried out using two public datasets: the WEEE dataset [27] and the cardiorespiratory measurement from graded cycloergometer exercise testing [28]. We use the abbreviation of the laboratory for the cardiorespiratory measurement, ACTES, to name the second dataset. 25 participants' data, aged between 12 to 40 (median: 18), from both datasets were applied in our study.

The WEEE dataset is a multi-device and multi-modal dataset collected from 17 healthy participants during different physical activities. Individual's oxygen uptake ($\dot{V}O_2$) data collected by $\dot{V}O_2$ Master Analyzer, and heart rate (HR) data collected by the Wahoo Tickr chest strap during physical activities were used for the PMB-NN model's development. The participants followed a physical activity protocol which consists of three types of physical activities: resting, cycling and running. During the resting phase, participants were instructed to sit on a chair for a duration of 5 min, followed by a subsequent 5-minute period of standing. Subsequently, participants engaged in aerobic activities, which were 10-minute cycling on an indoor stationary bike and 10-minute running on a treadmill. Both cycling and running activities were executed at two distinct intensity levels chosen by the individuals based on their fitness level [27], and each level activity lasted 5 min. The WEEE $\dot{V}O_2$ and HR data were synchronized to the same time scale and resampled with the sampling rate of 1 Hz in [27]. For our study, five participants' data were excluded from the WEEE dataset after our data screening according to theoretical physiological responses (n=3; 1/3 of the HR data from participants 07, 10, and 11 are inversely proportional to $\dot{V}O_2$ signals) and data missing (n=2; $\dot{V}O_2$ data after standing is missing from participant 14, and $\dot{V}O_2$ data for running is missing from participant 16). The applied participants' data was named as WEEE + number in the whole manuscript.

The ACTES dataset consists of cardiorespiratory measurements acquired from 18 adolescent athletes during cycloergometer maximal graded exercise tests. Beat-to-beat R–R interval data was derived from electrocardiogram (ECG) recordings from Cardio 110BT and $\dot{V}O_2$ data was collected by a Metalyzer 3B gas analyzer system. The participants followed a protocol consists of 3–4 minutes' preparation, 3 minutes' 50 watts cycling, incremental power cycling testing of 15 watts each minute until exhaustion, and 3 minutes' cycloergometer sitting recovery. Power of each period was recorded as 0 for preparation and recovery, value of watts for varies intensity cycling. For our study, five participants' data were excluded from the ACTES dataset after our data screening to partly data missing (number: 6, 8, 11, 16, 17). The applied participants' data was named as ACTES + number in the whole manuscript. From the current cohorts (WEEE + ACTES), the correlation between HR and $\dot{V}O_2$ spans from resting conditions through submaximal to near-maximal exercise. Specifically, HR ranged from approximately 60 to 200 bpm and $\dot{V}O_2$ from 0.2 to 3.8 L/min.

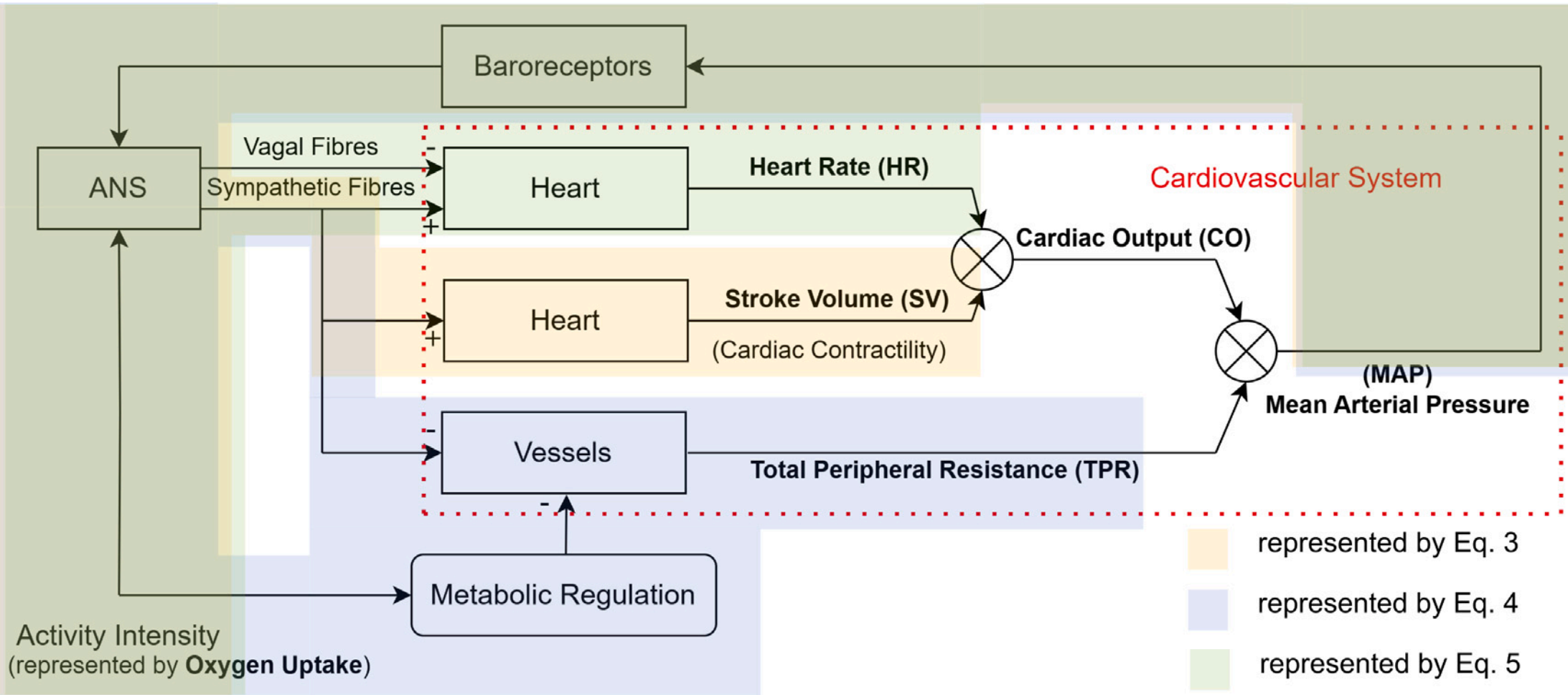

**Fig. 1.** Conceptual model of regulation of heart rate response to physical activity intensity. The bolded variables correspond to those included in the mathematical model outlined later in Section 2.2.2. The regulatory mechanisms, indicated by distinct colors in the figure, are detailed in Eqs. (3) to (5).

The 1 Hz sampled synchronized data was provided from WEEE dataset. For ACTES dataset, we calculated HR from the provided R–R interval, and resampled the synchronized $\dot{V}O_2$ and HR to 1 Hz via linear interpolation according to [29]. Given that $\dot{V}O_2$ signals are normally noisy due to large flow variation and motion artifacts especially in high-intensity activities [30], a first-order Savitzky–Golay high-pass filter [31] was applied to remove peak variations and motion artifacts in $\dot{V}O_2$ signals, and a low-pass finite impulse response filter [32] with window size of 10 samples was applied to smoothen both $\dot{V}O_2$ and HR signals, as this size is optimal for maintaining $\dot{V}O_2$ signal integrity while effectively removing noise. The signals before and after pre-processing are shown in Appendix A.

## 2.2. Physiological model

### 2.2.1. Conceptual model

Our proposed conceptual model describes a simplified close-loop heart rate control system responding to physical activity. A block diagram describing the regulatory mechanisms is shown in Fig. 1. This model mainly comprises: the cardiovascular system in 3 main compartments (heart mechanics, heart rate and vessels); a feedback pathway that represent the negative control loop of the baroreflex; a metabolic control block that takes into account the intensity of physical activity. This latter is modeled as a primary stimulus that induces changes in the metabolic regulation and autonomic nervous system (ANS). ANS has a direct effect on sympathetic activities affecting the heart and vessels, and on vagal activity affecting the heart rhythm only [13]. Sympathetic activities towards the heart result in cardiac contractility enhancement and increase in heart rhythm to increase cardiac output (CO). Both efferent sympathetic fibers on vessels and metabolic regulation cause total peripheral resistance (TPR) decrease which is essential to guarantee a higher perfusion to the exercising cardiovascular regions [13].

### 2.2.2. Mathematical model

In this study, we proposed a simplified physiological model (referred as PM throughout the manuscript) to estimate HR and incorporate physiological constraint in PMB-NN structure. We treated oxygen uptake ($\dot{V}O_2$; unit: L/min) and heart rate (HR; bpm) as our model input and output, respectively. Crucially, $\dot{V}O_2$ was selected as the model input,an indispensable intermediate variable from movement to HR change because it serves as the metabolic driver that causally regulates cardiovascular responses [33–35]. Unlike direct motion-to-HR mappings, using measured $\dot{V}O_2$ allows us to ground the estimation in the established cardiovascular–metabolic coupling mechanisms. The physiologically dynamic relationship between $\dot{V}O_2$ and HR change, which is illustrated in the conceptual model (Fig. 1), was simplified based on the following two assumptions and the classic circulation model from [13] for various freely physical activity occasions to focus on the key physiological mechanisms and reduce computational load for potential daily monitoring applications. Our two assumptions are listed below:

1. we assumed that the relationship between stroke volume (SV) and $\dot{V}O_2$, and between TPR and $\dot{V}O_2$ can be described as a logarithmic function, respectively, based on a previous study about the cardiovascular response of healthy adults during rest and graded intensity of cycling exercises [36].
2. we assumed that the HR is linearly related to the mean arterial pressure (MAP) changes during different activity intensities based on the approximately linear relationship between the change of HR and the change of MAP from rest to exercise observed in simulated normal subjects [13].

Explanation and validation procedure of the assumption is provided in Appendices B and C.

Accordingly, we mathematically described the simplified model as the following four algebraic equations: Eqs. (1)–(4) and a differential equation (DE): Eq. (5).

The cardiovascular system component illustrated in Fig. 1 was represented by the Eqs. (1)–(2). Cardiac output (CO) is the total volume of blood pumped by the ventricle per minute and defined as the product of the heart rate and the stroke volume, which is,

$$CO(t) = HR(t) \cdot SV(t) \tag{1}$$

where CO(t), HR(t) and SV(t) denote the time series of cardiac output, heart rate and stroke volume with the unit of L/min, beats/min, and L/beat, respectively. Mean arterial pressure (MAP) is the average arterial pressure throughout one cardiac cycle including the systole, and diastole. The relationship between the total peripheral resistance (TPR) and MAP was described by Eq. (2):

$$MAP(t) = CO(t) \cdot TPR(t) \tag{2}$$

where MAP(t) and TPR(t) denote the time series of mean arterial blood pressure with the unit of mmHg and total peripheral resistance with the unit of mmHg/L/min, respectively.

**Table 1**
Initial values and ranges for $\lambda_s$.

| $\lambda_s$ | Initial value | Explainable range | Unit |
|---|---|---|---|
| $\lambda_1$ | 0.02 | (0.01, 0.03) | L/min |
| $\lambda_2$ | 0.1 | (0.06, 0.15) | L/min |
| $\lambda_3$ | −5.3 | (−6, −2) | mmHg/L/min |
| $\lambda_4$ | 10.5 | (7, 20) | mmHg/L/min |
| $\lambda_5$ | 0.44 | (0.1, 0.6) | bpm/mmHg |
| $\lambda_6$ | 0.3 | (−0.5, 0.5) | bpm/min |

The logarithm functions to describe the relationship between SV and $\dot{V}O_2$, and between TPR and $\dot{V}O_2$ are described by the Eqs. (3) and (4), respectively.

$$SV(t) = \lambda_1 \cdot \ln \dot{V}O_2(t) + \lambda_2 \tag{3}$$

where $\lambda_1$ and $\lambda_2$, both with the unit of L/min, denote the unknown slope and bias parameters of the logarithmic function from $\dot{V}O_2$ to SV. $\lambda_1$ reflects the sensitivity of SV to changes in $\dot{V}O_2$, which indicates the heart's adaptive capacity to rising metabolic demand through augmentation of contractility and baroreflex control of autonomic outflow. $\lambda_2$ reflects the baseline offset of SV at a given reference $\dot{V}O_2$ which represents the intrinsic pumping capacity of the heart under resting conditions.

$$TPR(t) = \lambda_3 \cdot \ln \dot{V}O_2(t) + \lambda_4 \tag{4}$$

where $\lambda_3$ and $\lambda_4$, both with the unit of mmHg/L/min, denote the unknown slope and bias parameters of the logarithmic function from $\dot{V}O_2$ to TPR. $\lambda_3$ reflects the sensitivity of vascular resistance to changes in $\dot{V}O_2$, which captures the dynamic responsiveness of the vasculature to metabolic demand. $\lambda_4$ sets the baseline TPR level at a given reference $\dot{V}O_2$. The Appendix B provided detailed explanation and validation procedure of the assumption.

The linear relationship to represent the relationship between change of HR and change of MAP is described by the Eq. (5).

$$\frac{\partial HR(t)}{\partial t} = \lambda_5 \cdot \frac{\partial MAP(t)}{\partial t} + \lambda_6 \tag{5}$$

where $\lambda_5$ and $\lambda_6$, with the unit of bpm/mmHg and bpm/min, respectively, denote the slope and bias of the linear function between change of MAP and change of HR, respectively. $\lambda_5$ represents baroreflex sensitivity as the responsiveness of cardiac chronotropy to blood pressure fluctuations. $\lambda_6$ reflects the intrinsic baseline adjustment of HR which is not directly coupled to arterial pressure.

In summary, we solved the system of equations (Eqs. (1)–(5)) by substitution and derived the Eq. (6).

$$\frac{\partial HR(t)}{\partial t} = \lambda_5 \cdot \frac{\partial \left( HR(t) \cdot (\lambda_1 \log(\dot{V}O_2(t)) + \lambda_2) \cdot (\lambda_3 \log(\dot{V}O_2(t)) + \lambda_4) \right)}{\partial t} + \lambda_6 \tag{6}$$

A general first order differential equation form can replace Eq. (6):

$$HR_t = f(HR, \dot{V}O_2, t, \lambda s) \tag{7}$$

where $HR_t$ is the time derivative of HR and $f$ is the function of time derivative of product that includes HR and logarithmic of $\dot{V}O_2$ in Eq. (6). $\lambda_s$ represent $\lambda_1$ to $\lambda_6$ in Eq. (6). The data-fitted initial value and physiologically explainable ranges of $\lambda_s$ are defined based on normal healthy values in cardiovascular hemodynamics for the system identification and presented in Table 1, with detailed explanations available in Appendices B and C.

The performance of the sole physiological model was evaluated through fitting the mathematical formulation in Eq. (6), where the $\lambda_s$ were optimized via the limited-memory Broyden–Fletcher–Goldfarb–Shanno (LM-BFGS) algorithm on the training set and subsequently validated on the testing set for each individual subject.

### 2.3. PMB-NN architecture and algorithms

The structure of our PMB-NN framework is shown in Fig. 2. The PMB-NN was trained to respect both the measured HR training data fitting and the cardiac physiology mechanisms, which are described by our mathematical model in Section 2.2.2. Notably, the PMB-NN is built on the architecture of a fully connected neural network (FCNN), with an additional loss term for physiologically explainable estimation. The data fitting aspect was introduced to the PMB-NN framework via the data fitting term defined by a $L_2$ loss (cost) function ($L_{data}$) in the Eq. (8), and the physiology mechanism aspect was introduced via the physiological constraint term defined by a $L_2$ loss (cost) function ($L_{DE}$) in the Eq. (9).

$$L_{data} = \frac{1}{N} \sum_{i=1}^{N} (HR_{data,i} - HR_{pred,i})^2 \tag{8}$$

$$L_{DE} = \frac{1}{N} \sum_{i=1}^{N} (HR_{t,pred,i} - f(HR_{pred,i}, VO_{2pred,i}, t_i, \lambda))^2 \tag{9}$$

where $N$ is the number of data points, while $t_i$ is the time of each data point. $HR_{data,i}$ and $\dot{V}O_{2,data,i}$ denote the measured HR and $\dot{V}O_2$ value at $t_i$, respectively. $HR_{pred,i}$ and $HR_{t,pred,i}$ denote the estimated HR value and its time derivative at the same time of $t_i$. The unknown parameters in the mathematical model ($\lambda_1$ to $\lambda_6$) were assigned initial values and constrained within the ranges specified in Table 1.

Ultimately, we define the total loss $L_{tot}$ as:

$$L_{tot} = L_{data} + L_{DE} \tag{10}$$

We assign a weight of 1 to $L_{data}$ and $10^5$ to $L_{DE}$, ensuring that both loss terms operate on the same order of magnitude, thereby contributing equally to the optimization process of the PMB-NN framework.

Algorithm 1 gives a detailed flow illustration of the PMB-NN training algorithm.

**Algorithm 1:** PMB-NN Training Algorithm

**Require:** Training set containing $\dot{V}O_2$ and HR
1: **Initialization:**
2: Set neural network layer structure as 1-64-64-1
3: Set activation function as $Tanh$
4: Set the unknown parameters ($\lambda s$) in the mathematical model to initial values and constrain within the specific ranges
5: Set optimizer as RMSprop and learning rate as 0.01
6: **Loss terms definition:**
7: Define $L_{data}$ according to Eq. (8)
8: Define $L_{DE}$ according to Eq. (9)
9: Define $L_{tot} = L_{data} + L_{DE} \cdot 10^5$
10: **for** each epoch **do**
11: **Training Phase:**
12: Perform forward pass
13: Compute training loss ($L_{tot}$)
14: Perform back propagation and update model parameters including hidden parameters and $\lambda s$
15: Record training loss and $\lambda s$
16: **if** $L_{tot} < \epsilon$ **then**
17: **return** Hidden parameters and $\lambda s$ for testing phase
18: **else**
19: Repeat training for up to 5000 iterations
20: **return** Hidden parameters and $\lambda s$ for testing phase
21: **end if**
22: **end for**

The PMB-NN framework was trained on a workstation equipped with 12th Gen Intel(R) Core(TM) i7-12700H CPU. Deep learning framework Pytorch 2.1.0 in Python 3.9.13 was fully applied in establishing our algorithm. A fully-connected neural network structure was initialized. Two hidden layers and 64 neurons setting in each layer allowed for extraction and representation of complex features within the data and mitigated the risk of overfitting. Tangent activation function ($Tanh$) helped in learning nonlinear relationships between $\dot{V}O_2$ and HR therefore guaranteed the model's fitting capability. Setting the optimizer as Root Mean Squared Propagation (RMSprop) helped in maintaining a balance between speed and stability of the training process. The learning rate was chosen as 0.01 to empirically balance convergence speed and stability, ensuring efficient training of the model.

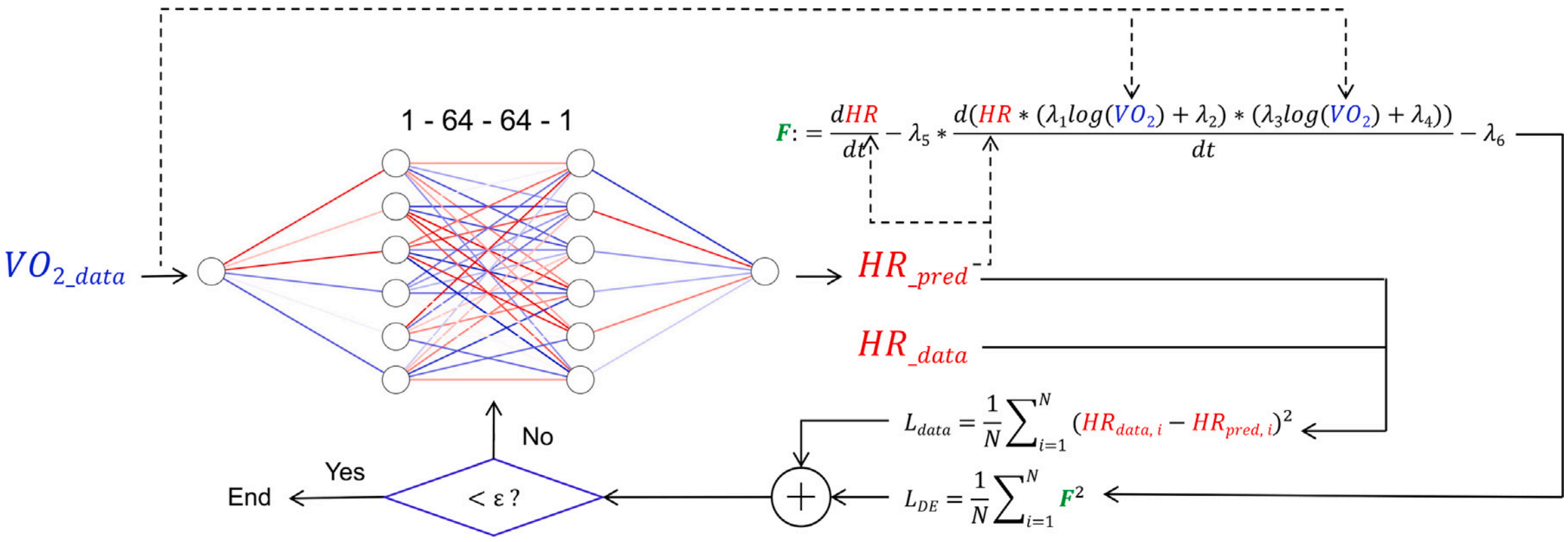

**Fig. 2.** Schematic of PMB-NN framework in this study. After initialization of the neural network structure and parameters in the physiological model, predicted HR ($HR_{pred}$) values were obtained from the inputs $\dot{V}O_2$ data by a forward pass multilayer perceptron (MLP). The $HR_{pred}$ values was substituted to both data fitting term ($L_{data}$) and the physiology constraint loss term ($L_{DE}$) afterwards. The sum of the two terms ($L_{tot}$) was used for random search to update both the physiological model parameters in Eq. (6) and the hyperparameters in MLP.

The PMB-NN framework was optimized by minimizing the loss function in Eq. (10), iteratively updating both the neural network weights and the unknown parameters. In each epoch, the model performed a forward pass, followed by backpropagation to calculate gradients, which were used in the weight update process via gradient descent. To ensure that physical constraints on model parameters were respected, the lambda parameters were initialized with values based on physiological knowledge and were updated during training. These parameters were mapped to a physically explainable range using a sigmoid function, preventing them from exceeding defined upper and lower boundaries. The training process was stopped once the total loss dropped below an empirically set threshold of 10 bpm or after 5000 epochs, whichever came first, revealing that the model had converged and further iterations would not significantly improve performance.

### 2.4. Dataset synthesization

Based on the pre-processed WEEE and ACTES datasets, we synthesized a new dataset that can capture the large activity variety nature of physical activities in daily living because the original datasets were collected with single non-repetitive prototype: resting-cycling-running for WEEE, preparing-graded cycling-recovering for ACTES. The dataset synthesization example process is illustrated in Fig. 3. In detail, the pre-processed WEEE dataset for each participant was divided into three segments according to the activity types: resting, cycling and running. Then each segment was divided into its 80% and 20%. The 80% of the three segments were merged into the training set, while the rest 20% segments were merged into the testing set. Similarly, the pre-processed ACTES data set for each participant was divided into multiple segments (the power corresponding to exhaustion is different for each participant, so the number of segments is different) according to the power value in original dataset. Same 80% : 20% division and merging procedure was conducted to create the training and testing set.

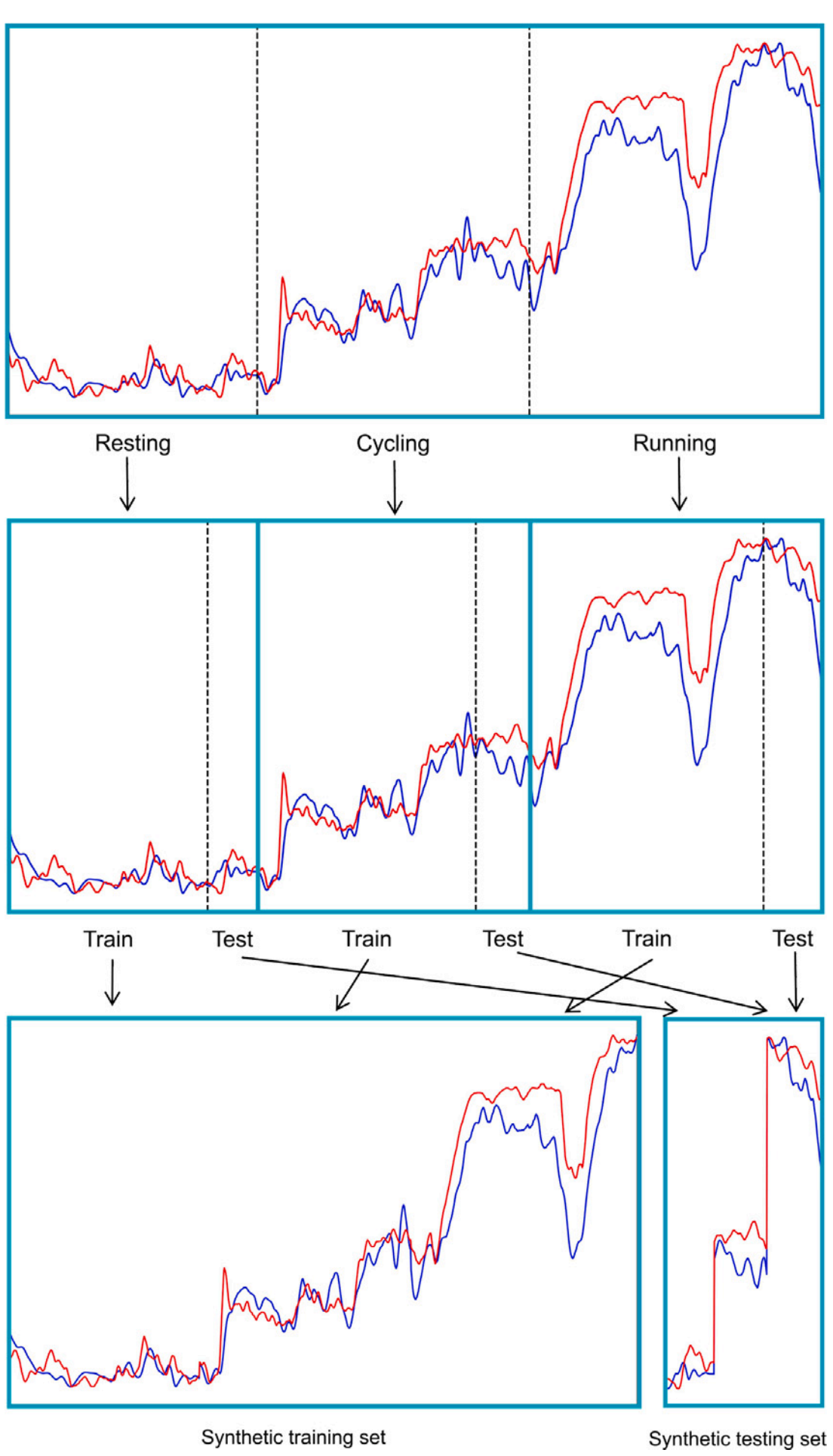

**Fig. 3.** Dataset synthesization illustration for WEEE dataset. The first row figure indicates the processed dataset including three activity types. The second row figures indicate the splitted segments according to activity types. Each segment was divided as training and testing set with the ratio of 8:2. The third row indicate the synthetic training set and synthetic testing set from splicing all the training and testing segments, respectively.

### 2.5. Model evaluation

We tested our four hypotheses in the following evaluation approaches:

1. To test the first hypothesis regarding the simplified PM model's capacity for HR estimation, we examined the performance of the PM model.
2. To test the second hypothesis, which posits that the PMB-NN outperforms the PM model, we compared the paired performance results of the PMB-NN and PM models.
3. To test the third hypothesis, which suggests that PMB-NN provides comparable estimation performance to the benchmark data-driven model and meanwhile improve its physiological plausibility, we conducted a paired comparison between the PMB-NN and FCNN models and validated the physiological coupling relationship between the model input $\dot{V}O_2$ and estimated HR through scatter plots and Spearman's correlation analysis.
4. To test the fourth hypothesis regarding PMB-NN's ability to identify physiologically explainable parameters, the personalized lambda values identified by PMB-NN were substituted into the PM model. This method, named as 'PMB-NN-R', resulted in the output, referred to as 'reconstructed HR'. We then compared the paired performances of the PMB-NN-R and PM models to validate the effectiveness of the lambda values identified by PMB-NN.

PMB-NN, FCNN and PM were trained and tested on the same training and testing sets described in Section 2.4, respectively. We evaluated each model's performance through evaluation metrics as follows:

1. The coefficient of determination is denoted by $R^2$ and indicates the proportion of the variance in measured HR that can be explained by the estimated HR. The $R^2$ is defined in the Eq. (11) where $HR_i$ and $\hat{HR}_i$ represent reference value and estimated value of HR, respectively. $\bar{HR}$ in the denominator represents the average value of $HR_i$ and $n$ denote the number of data points.
2. The root mean square error (RMSE) represents the standard deviation of the residuals and measures the average magnitude of the errors between the estimated HR and the reference HR. RMSE is defined in Eq. (12) where $HR_i$, $\hat{HR}_i$ and $n$ have the same definitions as mentioned in Eq. (12).

$$R^2 = 1 - \frac{\sum_{i=1}^{n}(HR_i - \hat{HR}_i)^2}{\sum_{i=1}^{n}(HR_i - \bar{HR})^2} \tag{11}$$

$$RMSE = \sqrt{\frac{1}{n}\sum_{i=1}^{n}(HR_i - \hat{HR}_i)^2} \tag{12}$$

The mean absolute error (MAE) between the estimated and the reference HR was calculated to provide reference comparing with state-of-the-art HR estimation studies. Median, maximum and minimum values of the above evaluation metrics and the computation time were also chosen to evaluate the model's overall performance. The median values were chosen to guarantee the robustness to outliers given the relatively small sample size (data from 25 participants). The maximum and minimum values were chosen to assess the range of model performance across different conditions. Boxplots were used to visually represent the distribution of the evaluation metrics, including the median, maximum, and minimum values, as well as the spread and potential outliers in the data. This provides a clear overview of the variability and robustness of the model's performance across participants.

To assess the physiological plausibility of the HR-$\dot{V}O_2$ coupling, we analyzed both qualitative and quantitative relationships between oxygen consumption and heart rate. For qualitative assessment, predicted HR from each model was plotted against $\dot{V}O_2$ for each participant. Quantitatively, Spearman's rank correlation coefficient ($\rho$) was computed between $\dot{V}O_2$ and predicted HR for each participant and model and summarized as median (min, max) across participants.

The Wilcoxon signed-rank tests [37] were performed on each pairwise comparisons of the above described metrics between PMB-NN and the respective models (25 pairs for 25 participants). A $p$-value from one-tailed comparison less than or equal to the significance level of 0.05 was treated as statistically significant difference. Cohen's d was calculated on each same pairwise comparisons to quantify the effect size, with values of below 0.2, between 0.2 and 0.5, between 0.5 and 0.8, and above 0.8 representing trivial, small, medium, and large effects, respectively.

### 2.6. Sensitivity analysis

To assess the generalizability of the proposed framework for future daily life applications where $\dot{V}O_2$ may be estimated from IMU sensors (introducing input uncertainty), sensitivity analysis with rigorous stress testing was conducted. We evaluated the robustness of the trained PMB-NN, FCNN, and PM models by injecting synthetic noise into the $\dot{V}O_2$ signals in the model testing phase. Three types of input uncertainty scenarios were simulated and described in the 2.6.1–2.6.3. For each uncertainty level, we repeated the evaluation and recorded the median $R^2$, RMSE and Spearman's $\rho$ degradation for all three models. Two-tailed statistical comparisons were performed between the PMB-NN and the two baseline models for each metric across all uncertainty scenarios, where p and d values were recorded.

#### 2.6.1. Gaussian noise

We introduced multiplicative Gaussian noise to simulate continuous estimation errors inherent in sensor-derived data. The noisy input $\dot{V}O_{2,\text{noise}}(t)$ is defined as:

$$\dot{V}O_{2,noise}(t) = \dot{V}O_2(t) \cdot (1 + \epsilon(t)) \tag{13}$$

where $\epsilon(t) \sim \mathcal{N}(0, \sigma^2)$. We varied the noise level $\sigma$ from 5% to 50% with a step size of 5%. While typical estimation errors for wearable sensors generally fall within a lower range (e.g., 10%–20%), we intentionally extended the upper limit to 50% to simulate extreme "worst-case" scenarios of sensor degradation.

#### 2.6.2. Random outliers

To simulate sudden signal artifacts or motion disruptions common in wearable sensors, we introduced random outliers into the clean $\dot{V}O_2$ signal. Unlike Gaussian noise, these outliers represent distinct, short-duration distortion events. The outlier injection is defined as:

$$\dot{V}O_{2,outlier}(t) = \begin{cases} \dot{V}O_2(t) \cdot (1 + \delta) & \text{with probability } p \\ \dot{V}O_2(t) & \text{with probability } 1 - p \end{cases} \tag{14}$$

where $p$ represents the probability of an outlier event occurrence, ranging from 0.5% to 5.0% with a step size of 0.5%. When an outlier event is triggered, it persists for a window of 2 s (simulating a sustained artifact rather than a single point spike). The magnitude factor $\delta$ was randomly sampled to induce severe deviations, where $\delta \in \pm[0.5, 1.0]$, corresponding to an amplitude shift of ±50% to ±100%.

#### 2.6.3. Time lag

To simulate potential processing delays or synchronization mismatches between the sensor-derived input and the physiological response, we introduced temporal shifts. The lagged input is defined as:

$$\dot{V}O_{2,lag}(t) = \begin{cases} \dot{V}O_2(t - \tau) & \text{if } t > \tau \\ \dot{V}O_2(0) & \text{if } t \le \tau \end{cases} \tag{15}$$

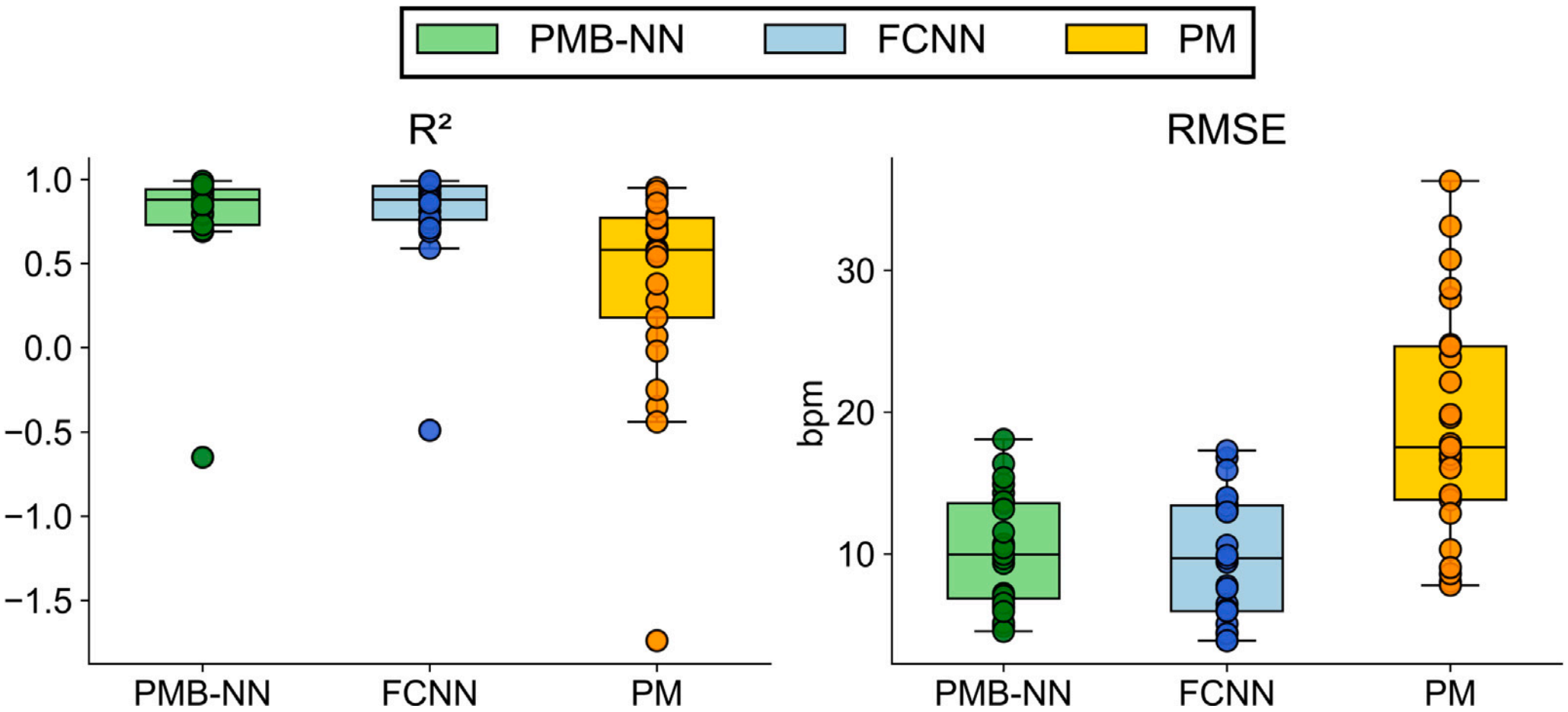

**Fig. 4.** Performance comparison between PMB-NN, FCNN frameworks and PM in HR estimation tasks using box plot. The subplots show the performance of each model across two evaluation metrics: $R^2$ and RMSE (bpm). In each subplot, the box plot displays the distribution of each metric, with the middle line indicating the median, the box representing the interquartile range (IQR), and the whiskers extending to 1.5 times the IQR or the minimum/maximum value within this range. Individual points outside the whiskers are considered outliers. Green represents the PMB-NN framework, blue represents the FCNN framework, while yellow represents the PM framework.

where $\tau$ represents the time lag, ranging from 2 s to 20 s with a step size of 2 s. This simulates scenarios where the estimated metabolic demand lags behind the actual physiological state due to sensor latency or algorithm processing time.

## 3. Results

### 3.1. Performance evaluation

The overall estimation performance of the PM, PMB-NN and FCNN models were summarized in the following. Both PMB-NN and FCNN reached an appropriate estimation accuracy respectively with $R^2$ of 0.88 (−0.65, 0.99) and 0.88 (−0.49, 0.99), RMSE of 9.96 (4.54, 18.08) bpm and 9.69 (3.88, 17.30) bpm, while the PM model yielded a relatively lower $R^2$ of 0.58 (−1.74, 0.95), RMSE of 17.53 (7.8, 36.30) bpm. Median MAE for PMB-NN is 8.87 bpm.

The performance of PMB-NN, FCNN, and PM on individual participants is given in Table 2, Appendix D. The Wilcoxon signed-rank test revealed that PMB-NN provided significant improved accuracy comparing with PM reflecting in both $R^2$ ($p < 0.001$, d = −0.832) and RMSE ($p < 0.001$, d = 1.398), meanwhile reached similar accuracy comparing with FCNN reflected as $R^2$ (p = 0.305, d = 0.018) and RMSE (p = 0.059, d = −0.075). Fig. 4 provided the performance summary of both PMB-NN, FCNN, and PM and comparison by listing each evaluation metric in Table 2. A detailed statistical comparison of the three models' performance on WEEE dataset, evaluated across distinct activity types, is presented in Appendix E. Activity-specific performance revealed that PMB-NN outperformed PM in cycling (RMSE: p = 0.034, d = 0.686). No significant differences were found between PMB-NN and FCNN across all activities.

We presented the PMB-NN, FCNN, and PM results of four representative participants as visualization examples in Fig. 5. In WEEE-01, all three models demonstrated good agreement with the real values during the resting and cycling phase, while PMB-NN and FCNN exhibited relatively lower errors than PM in the running phases. In WEEE-03, PM exhibited better HR alignment during running phase. For both ACTES-2 and ACTES-12, HR estimation from PMB-NN and FCNN revealed closer alignment than PM during all phases.

The HR-$\dot{V}O_2$ coupling of three models are visually represented in Fig. 6. Specifically, the PMB-NN model closely follows the physiological trend with a smooth monotonic increase in HR, whereas the FCNN model displays a characteristic zig-zag pattern. The PM model generally captures the smooth monotonic relationship between HR and $\dot{V}O_2$. However, due to its continuous recursive nature, the PM model sometimes struggles with accuracy. Specifically, in certain cases, the PM model predicts HR to increase too slowly as $\dot{V}O_2$ rises or conversely, it predicts HR to increase too rapidly, exceeding the expected submaximal HR range. Detailed values of Spearman's $\rho$ between $\dot{V}O_2$ and estimated HR for each participant, along with a comparison among the three models, are provided in Table 3 in Appendix D. The median (min, max) values of $\rho$ for the PMB-NN, FCNN, and PM models are 1 (0.999, 1), 1 (0.863, 1), and 0.996 (−0.162, 1), respectively. Statistical analysis reveals that PMB-NN outperforms both FCNN (p = 0.028, d = −0.372) and PM ($p < 0.001$, d = −0.276) in capturing the HR-$\dot{V}O_2$ coupling.

Table 4 in Appendix E presents the PMB-NN-identified personalized $\lambda$ values and their corresponding reconstruction (PMB-NN-R) performance metrics, and Table 5 in Appendix E presents the PM identified $\lambda$ values. The results indicate that PMB-NN offers greater flexibility and variability compared to PM. While PM maintains a more consistent set of lambda values across participants, PMB-NN shows a broader range of values. There was no significant performance difference between PMB-NN-R ($R^2$: 0.66 (−0.34, 0.93); RMSE: 15.80 (7.56, 35.23)) and PM estimation (comparison: $R^2$: p = 0.507, d = −0.227; RMSE: p = 0.672, d = 0.072).

### 3.2. Sensitivity analysis

The sensitivity analysis reveals that the PMB-NN sustains an RMSE within 15 bpm when subjected to input perturbations of up to 20% Gaussian noise, 4% random outliers, and an 18 s temporal lag. Meanwhile PMB-NN achieves statistically equivalent numerical accuracy to FCNN and higher accuracy to PM across all uncertainty scenarios. Crucially, the PMB-NN consistently provides superior physiological plausibility than both baseline models under uncertainty conditions, validating the effectiveness of the physics-informed loss term in preventing non-physiological prediction artifacts. Fig. 7 illustrates the performance degradation of the PMB-NN, FCNN, and PM models under three noise scenarios: Gaussian noise, random outliers, and time lag.

Robustness against Gaussian noise: As the intensity of Gaussian noise ($\sigma$) increased from 0% to 50%, the estimation error for all models rose while $R^2$ declined. The PMB-NN and FCNN exhibited highly similar degradation patterns. Pairwise statistical analysis of the median performance curves revealed no significant difference between PMB-NN and FCNN in terms of RMSE (p = 0.831, d = 0.054) or $R^2$ (p = 0.074, d

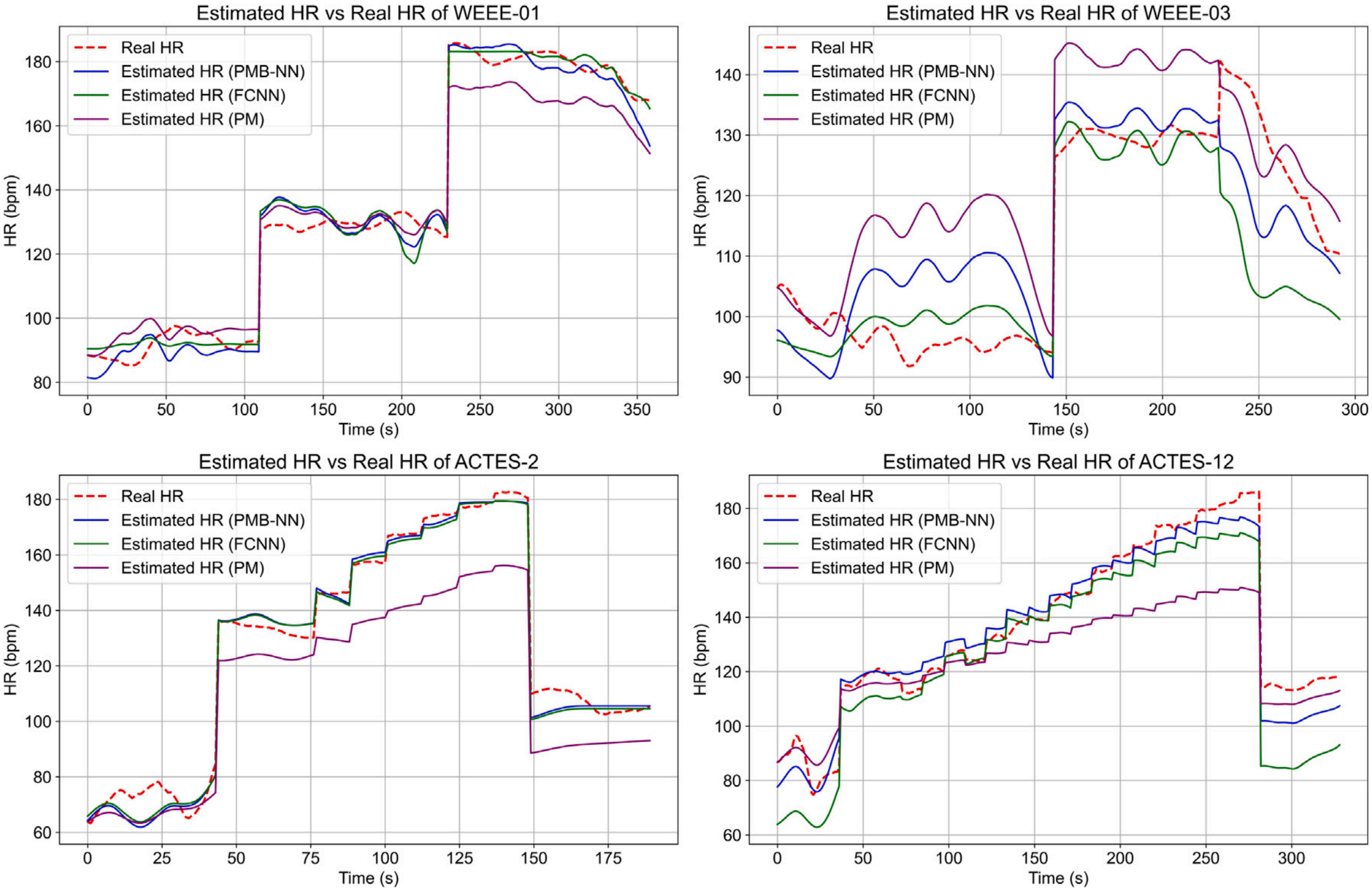

**Fig. 5.** Three models' HR estimation results of participant WEEE-01, WEEE-03, ACTES-02, ACTES-12. PMB-NN, FCNN, PM estimated HR comparing with real HR are presented in each subfigure. The red dash curve in each plot represents the measured HR values, while the blue, green and purple solid curves represent the PMB-NN, FCNN and PM estimated HR values, respectively.

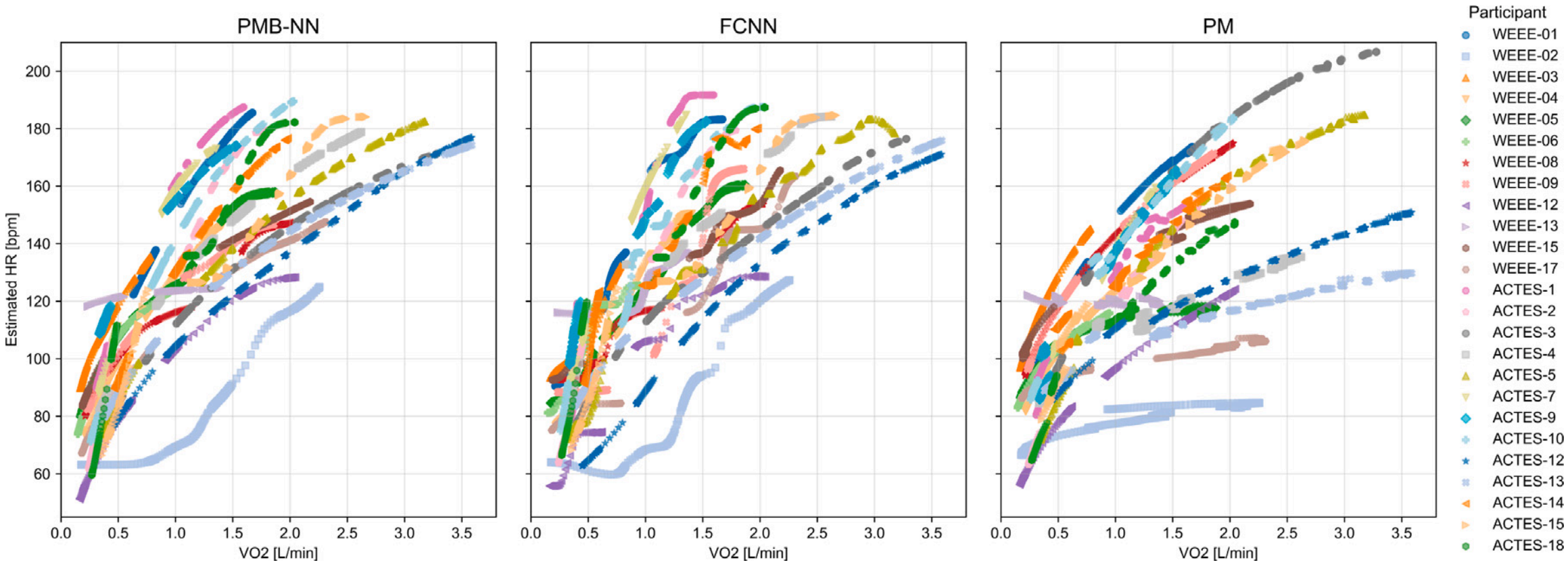

**Fig. 6.** Scatter plot for $\dot{V}O_2$ and estimated HR for each participant from three models. Subfigures from left to right indicate results from PMB-NN, FCNN and PM, respectively. Points in different color and shape refers to results from different participants.

= 0.556). However, a distinct divergence occurred in the physiological coupling metric. As shown in the bottom row of Fig. 7, the FCNN's ability to maintain monotonic coupling collapsed when noise exceeded 30%, whereas the PMB-NN maintained a robust median correlation of $\rho = 1.0$. Consequently, PMB-NN significantly outperformed FCNN in physiological consistency ($p < 0.001$, d = −1.001). The PM model showed the lowest sensitivity to noise but consistently performed worse than PMB-NN in estimation accuracy ($p < 0.001$).

Robustness against random outliers: Under the influence of random outliers (p = 0% to 5%), PMB-NN demonstrated comparable accuracy with FCNN, while significantly outperform PM and uniquely maintained perfect physiological plausibility. Both PMB-NN and FCNN maintained relatively high $R^2$ scores (> 0.7) and lower RMSE (< 16 bpm) compared to the PM model, which exhibited markedly lower accuracy ($R^2$ between 0.4 and 0.6, RMSE > 17.5 bpm). Statistical tests revealed no significant difference between PMB-NN and FCNN for RMSE (p = 0.413, d = 0.289) or $R^2$ (p = 0.074, d = 0.667). However, the PMB-NN achieved a perfect median Spearman's $\rho$ of 1.0, significantly outperforming both the FCNN ($p < 0.001$, d = −1.994) and the PM model ($p < 0.001$, d = −5.538), demonstrating its superior ability to preserve the directional relationship between metabolic input and cardiac response during artifact events.

Robustness against time lag: PMB-NN maintained perfect physiological plausibility ($\rho = 1.0$) across all time lag intervals, even as absolute predictive accuracy degrades for all models. While the PM model showed a relatively stable but consistently lower baseline accuracy ($R^2$ of 0.6), both neural networks demonstrated superior performance at lower lag intervals before converging with the PM baseline at extreme lags ($\tau > 18$ s). Comparison between the PMB-NN and FCNN showed no statistically significant difference for RMSE (p = 0.700, d = −0.180) or $R^2$ (p = 0.059, d = −0.538), indicating a similar susceptibility to temporal misalignment in terms of numerical error. Crucially, the PMB-NN significantly outperformed both FCNN ($p < 0.001$, d = −2.569) and the PM model ($p < 0.001$, d = −1.009) in Spearman's $\rho$, confirming that its embedded constraints effectively safeguard output plausibility when temporal synchronization is compromised.

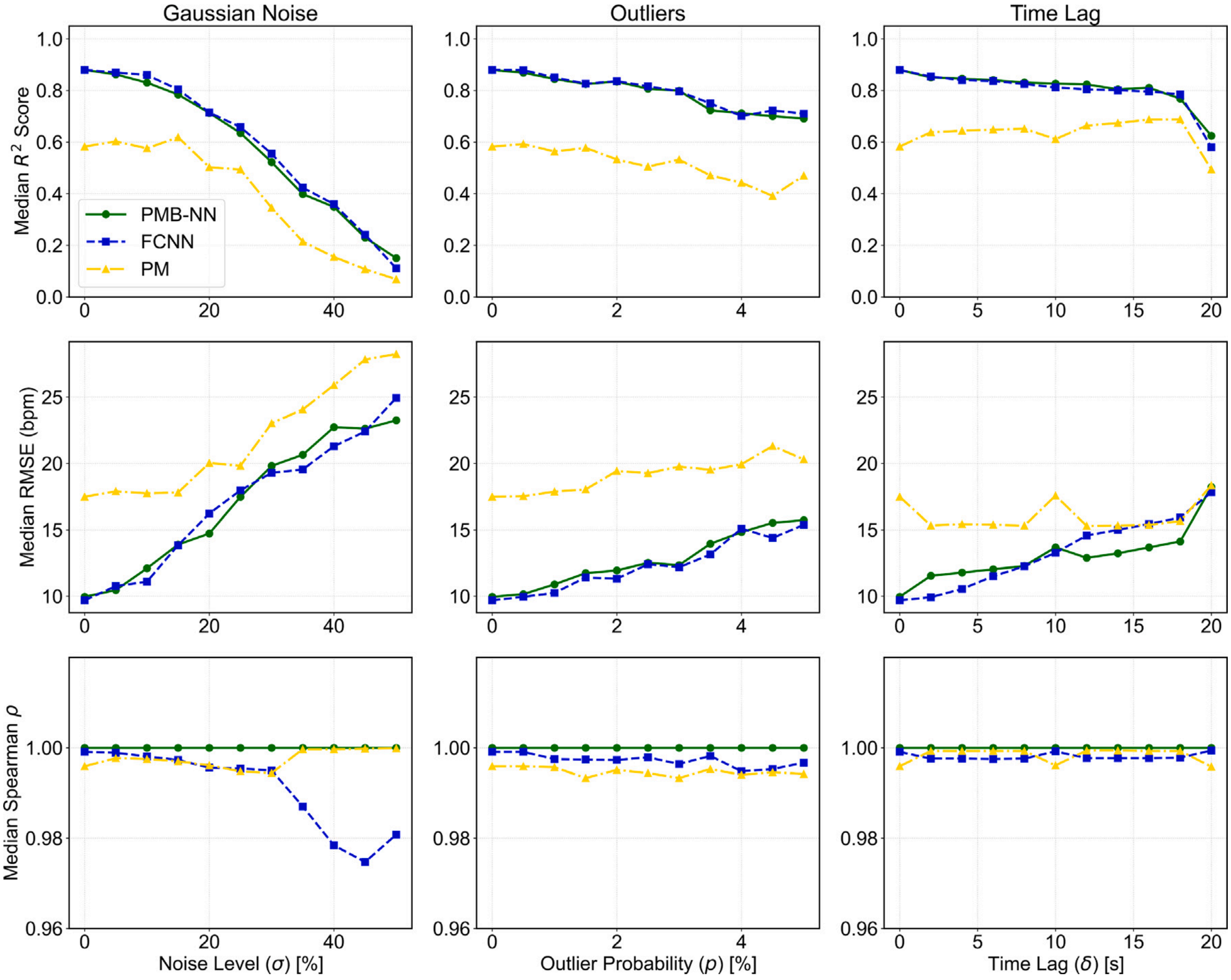

**Fig. 7.** Sensitivity analysis of HR estimation models under three input uncertainty scenarios. The rows represent the median values of the Coefficient of Determination ($R^2$, top), Root Mean Square Error (RMSE, middle), and Spearman's Rank Correlation Coefficient ($\rho$, bottom) across the testing dataset. The columns correspond to the three simulated noise types: (Left) Gaussian noise with intensity $\sigma$ ranging from 0% to 50%; (Middle) random outliers with probability $p$ ranging from 0% to 5%; and (Right) time lag $\delta$ ranging from 0 to 20 s. The green solid line represents the proposed PMB-NN, the blue dashed line represents the benchmark FCNN, and the yellow dot-dashed line represents the physiological model (PM).

## 4. Discussion

This study proposes a physiological model-based neural network structure for heart rate estimation from oxygen uptake signal. The primary contribution of this work lies in the successful integration of physiological principles into a data-driven neural network, which not only enhances the interpretability of the model but also achieves excellent estimation accuracy. The results of this study validate our initial hypotheses. Firstly, the simplified PM is able to demonstrate a reasonable capability in HR estimation. Next, the PMB-NN model significantly outperforms the conventional physiological model meanwhile achieves equivalent performance to the benchmark fully connected neural network model. With an overall $R^2$ of 0.88 and RMSE of 9.96 bpm, the PMB-NN framework demonstrated excellent accuracy ($R^2 \geq 0.75$) [38] in HR estimation. Moreover, PMB-NN most accurately captures the expected physiological HR-$\dot{V}O_2$ coupling. Finally, the PMB-NN is capable in physiologically relevant parameter identification. Deeper exploration of general performance and certain nuanced aspects of individual performance were discussed to understand the limitations, perspective of further development and potential applications of this model.

### 4.1. Heart rate estimation analysis

#### 4.1.1. Overall performance and physiological consistency

We observe noticeable performance improvement of PMB-NN compared with PM in Table 2. The superior performance of PMB-NN over PM can be attributed to its hybrid modeling architecture, which couples a fully connected neural network (FCNN) with an explicit physiological loss term. This design allows PMB-NN to overcome the limitations of traditional parametric models by capturing complex, non-linear relationships. The FCNN component allows the model to flexibly approximate a broad range of individual-specific mappings, while the physiological loss term introduces inductive bias grounded in domain knowledge, constraining the solution space towards physiologically explainable behaviors. This dual mechanism enhances both expressivity and interpretability: the network can fit participant-specific data patterns with high fidelity, while the physiological constraint ensures model estimations remain consistent with known physiological principles. Importantly, this synergy enables PMB-NN to generalize better across heterogeneous populations and varying conditions, ultimately yielding superior predictive accuracy and robustness. Additionally, PMB-NN excels at capturing the HR-$\dot{V}O_2$ coupling by ensuring a smooth, monotonic relationship between HR and $\dot{V}O_2$, which the FCNN struggles with due to its sensitivity to noise and fluctuations, and the PM fails to fully capture due to its limited flexibility. By combining the expressivity of FCNN with the physiological constraints of PM, PMB-NN is able to maintain physiological realism while still adapting to individual-specific patterns and providing accurate, interpretable predictions.

The PMB-NN achieves performance comparable to the FCNN in Table 2, by combining the expressive power of neural networks with physiological constraints that guide estimations towards biologically explainable solutions. This integration enables the model to capture individual-specific nonlinear patterns while avoiding overfitting, thus maintaining accuracy on par with the benchmark. Slight differences in $R^2$ ($\geq 0.04$) are observed for 9 out of 25 participants, based on pairwise estimation plots and the corresponding metrics in Table 2. These discrepancies may arise from the increased optimization complexity introduced by the physiological constraints. Analysis of training loss curves revealed that, compared to the relatively stable data loss, the physiological loss exhibited substantial fluctuations, indicating that the model struggled to consistently satisfy the differential equation constraint. This reflects an inherent trade-off between minimizing estimation error and enforcing physiological plausibility. Moreover, the limited dataset and invariant structure may have further restricted model performance. To create sufficient training input, we segmented and concatenated the data into shorter windows, a process that likely introduced noise and disrupted temporal continuity of physiological signals. From the interpretable AI perspective, the model's reliance on physiological constraints introduces a layer of transparency by ensuring that the estimated parameters adhere to known biological principles. This provides valuable insight into how the model's internal representations align with real-world physiological processes, which makes it more interpretable and fostering trust in the model's outputs. Future work could collect more diverse and continuous datasets, or adopt augmentation techniques that preserve physiological structure. These may enhance model generalizability across different exercise types and participants.

The prevalence of negative $R^2$ values in Appendix F underscore the metric's susceptibility to model misspecification, as it penalizes deviations from an oversimplified constant mean baseline, whereas RMSE retains interpretability by quantifying absolute estimation errors independent of reference models. The integration of p-values and Cohen's d effect sizes becomes critically instructive in our stratified analyses, where both limited sample size ($n = 25$) and truncated data segments per activity synergistically constrain statistical power. While conventional p-values struggle to detect significance in fragmented datasets, as observed in the resting-state comparisons where PMB-NN exhibited a moderate effect ($d = -0.607$) in RMSE but non-significant underperforming p-values ($p = 0.064$) comparing with FCNN, effect sizes quantify the magnitude of differences independent of sample variability, unmasking trends obscured by motion-specific data partitioning. For instance, in cycling, PMB-NN's RMSE advantage to PM achieved both statistical significance ($p = 0.034$) and a robust effect ($d = 0.686$), whereas in running, yielding negligible effects ($d = 0.548$) despite comparable p-values ($p = 0.077$). These results suggest that with a larger sample size, the potential performance advantage of PMB-NN over traditional models may become more pronounced, helping to detect more subtle yet important trends that are currently masked by the smaller sample size. This duality underscores that effect sizes contextualize p-values by distinguishing statistical noise from biologically explainable effects, particularly when activity stratification reduces within-group data continuity.

It is noted that the performance of PM and PMB-NN demonstrated a consistent pattern across different participants; both models excelled or underperformed in tandem. This synchronicity implies that the inherent characteristics of individual participants exert a uniform influence on the efficacy of both models. An example of processed data from WEEE-02 is given in Appendix H. Upon closer inspection of the processed data for eight participants (WEEE-02, 03, 06, 09, 15 and ACTES-3, 4, 5), a shared trend emerged: each exhibits a sudden increase in HR after first 5 min for WEEE and surrounding the third minute for ACTES, from approximately 60–70 bpm to 90–100 bpm, which can be attributed to a specific physiological response to the "sit-to-stand" stimulus. This stimulus activates the sympathetic nervous system, prompting an immediate rise in HR as the body adjusts to postural change. Following this initial surge, HR gradually decreases to its resting level. However, during this HR adjustment period, $\dot{V}O_2$ remains relatively stable, showing minimal fluctuations. This lack of proportionality between the HR and $\dot{V}O_2$ signals creates a unique challenge for the PMB-NN model, which relies on the assumed relationship (described in the Eqs. (3)–(5)) between these two physiological indicators.

On the other hand, WEEE-13 demonstrates an relatively low $R^2$ of $-0.65$ of PMB-NN. According to the original data set of this participant, $\dot{V}O_2$ showed a rapid increase at the onset of cycling and stays in the same level during the first 5 min of cycling. However, HR paradoxically decreased from 130 to 100 bpm within the same period. This response, while unusual, is not entirely unprecedented. In some rare cases, a decrease in HR during the onset of exercise can be attributed to a phenomenon known as exercise-induced bradycardia or a vagal response, where the parasympathetic nervous system remains dominant during initial physical exertion [39]. This could result in a temporary drop in HR despite the increase in $\dot{V}O_2$, as the body adapts to the onset of exercise [39]. Additionally, such responses might also be observed in individuals with high cardiovascular fitness levels, where the heart becomes more efficient in pumping blood and may initially decrease in rate before stabilizing or increasing with prolonged activity [40]. Retaining this participant's data within the model input allows us to better assess the model's generalization capacity across atypical cases. The reliability of the PMB-NN was demonstrated by its strong performance. However, the unsatisfactory performance on WEEE-13 indicates that the model's ability to generalize across a diverse range of data types still needs improvement. Future extensions could integrate beat-to-beat blood pressure estimated from PPG signals, which provide information about baroreflex-mediated heart rate regulation, as additional input. Such multimodal input may improve prediction accuracy and reduce errors in participants with unusual physiological responses, thereby enhancing the generalizability of the PMB-NN framework.

Although our framework requires an initial calibration phase to train a personalized model for each individual, this step is performed only offline. Crucially, the complex integration of physiological constraints is restricted to this training phase via the loss function. During the real-time application, the trained PMB-NN functions as a standard, lightweight fully connected neural network without the need to numerically solve equations. Consequently, it demonstrates exceptional computational efficiency, requiring only approximately 1 MFLOPs per inference. This low computational cost enables the model to generate HR predictions in real-time with minimal latency and power consumption, making it highly suitable for deployment on resource-constrained wearable devices. The model's lightweight nature (requiring less than 20KB of storage for parameters) further mitigates memory and storage constraints commonly encountered in wearable devices. Nevertheless, to address potential deployment challenges in resource-constrained environments, additional optimization strategies including model compression, weight quantization, knowledge distillation, and deployment via lightweight edge-computing frameworks may be considered for further efficiency gains.

in Appendix I provides a summary of representative state-of-the-art HR estimation studies, serving as contextual reference given the diversity of datasets, input modalities, and testing protocols across works. Unlike most prior studies that rely on relevant temporal HR information, the present framework estimates HR from the metabolic consumption $\dot{V}O_2$, enabling estimation under highly dynamic exercise conditions. This inherently increases modeling difficulty, which is reflected in the median MAE of 8.87 bpm, higher than values typically reported for static or correlation-based approaches. Nevertheless, PMB-NN captured individualized HR–metabolic relationships with an $R^2$ of 0.88, demonstrating its capacity to reproduce physiologically coherent response dynamics. Compared with our earlier data-driven model [23], which achieved an $R^2 = 0.81$ and MAE = 6 bpm using body-movement intensity as input, PMB-NN achieves improved robustness and interpretability across a broader activity spectrum. These results suggest that incorporating metabolic dynamics through physiology–model coupling enhances the model's generalizability and provides a foundation for personalized HR digital-twin development.

#### 4.1.2. Robustness against input uncertainty and implications for daily life implementation

The results demonstrate that $\dot{V}O_2$ inputs corrupted by a 20% Gaussian noise level, alongside 4% random outliers and an 18 s time lag, lead to a median HR RMSE of approximately 15 bpm. While this error exceeds laboratory-grade precision, it defines a functional resolution limit for heart rate estimation in unconstrained environments. Beyond this 20% uncertainty threshold, prediction accuracy degrades markedly, establishing a practical robustness envelope for daily living applications.

This 15 bpm margin serves as a quantitative noise floor; deviations between measured and estimated HR occurring within this range are interpreted as modeling uncertainty. For long-term monitoring, this accuracy remains acceptable as the objective shifts from capturing instantaneous, beat-to-beat precision to identifying macroscopic response levels and temporal trends.

Furthermore, the approximately increase in RMSE observed for both PMB-NN and FCNN suggests that performance degradation is primarily constrained by the intrinsic quality of the $\dot{V}O_2$ estimates rather than architectural instability. Notably, the PMB-NN maintains behavioral consistency under noisy conditions, avoiding the erratic or physiologically impossible HR oscillations often seen in purely data-driven models. This property ensures that the digital twin provides a stable, biology-grounded reference, enabling the system to contextualize wearable-derived data effectively despite unavoidable sensor noise.

#### 4.1.3. Interpretation of heart rate deviations using a digital twin baseline

In practical deployment, the proposed framework functions as a real-time digital twin that operates in parallel with measured heart rate (HR) rather than replacing wearable sensing. By integrating estimated $\dot{V}O_2$ and learned personal dynamics, the digital twin generates an expected HR trajectory that provides a continuous physiological context for the observed data. Under this paradigm, the primary value of the system lies in the comparison between predicted and measured values; the model acts as a dynamic baseline that allows for the differentiation between expected variability and potentially abnormal trends.

To operationalize this comparison, a principled protocol for interpreting deviations must be established. The personalized model is first calibrated using an individual's healthy reference data, which defines a model-specific uncertainty or "noise floor". This margin accounts for the inherent noise introduced by both wearable-derived inputs and data-driven modeling. Consequently, any discrepancies occurring within this baseline threshold are interpreted as estimation variability rather than meaningful physiological change.

A true abnormal response is therefore inferred not from isolated, point-wise errors, but from systematic and sustained deviations that consistently exceed the established noise floor. This distinction is critical for clinical interpretability: while random errors are typically zero-mean and transient, physiological abnormalities manifest as structured divergences over time. For example, a persistent failure of the measured HR to reach the model-predicted target during periods of increased metabolic demand may indicate an impaired rate response [7]. Similarly, a systematic delay in the HR returning to estimated resting levels after exertion may reflect altered autonomic regulation [4,5].

To further enhance the robustness of this framework in real-world scenarios, future analyses could formalize these comparisons through time-series-based deviation metrics. Employing techniques such as cumulative error measures, trend consistency indices, or change-point detection methods would allow for a more systematic capture of sustained divergences, moving beyond simple thresholding towards a more sophisticated diagnostic digital twin.

### 4.2. Parameter identification analysis

The PMB-NN demonstrates several advantages over the traditional PM approach in parameter identification. PMB-NN is able to successfully identify individualized $\lambda$ values across participants. Unlike PM, where many parameter estimates clustered near the boundaries of the predefined search space, the PMB-NN produces a more diverse and physiologically explainable distribution of parameter values. This suggests that the data-driven learning mechanism in PMB-NN allowed for better adaptation to participant-specific features, avoiding the boundary effects that are often observed in optimization-based fitting procedures under rigid constraints. Moreover, PMB-NN achieve reconstruction accuracy comparable to PM, results for both specifically shown in Tables 2 and 4. This indicates that the model preserved the predictive strength of the original system while enabling more flexible and individualized parameter estimation. The physiological constraint, enforced through a differential equation-based loss term, likely contributed to the identifiability and realism of the learned parameters without compromising estimation quality. Among the performance metrics of PMB-NN-R, we observe a notable underperformance in WEEE-06, ACTES-9 and ACTES-15, where the $R^2$ value for WEEE-06 was 0.04, compared to 0.69 for the PM model, as shown in Table 2. Upon further investigation of the estimation curves for both models, presented in Appendix G, we speculate that the physiological constraints introduced by the PMB-NN model may have caused excessive misalignment for WEEE-06. The larger discrepancy could be due to the model's difficulty in accurately learning and generalizing the parameter values, particularly with limited data or the complexity of individual-specific cardiovascular dynamics. Also, this observation highlight the model's sensitivity to variations. Further tuning of the parameters or adjusting of the model could potentially address these discrepancies and improve the performance of PMB-NN-R in such cases.

Our findings also highlight the successful construction of a simplified physiological model (PM) that captures key individual-specific parameters using a compact set of interpretable variables. This model, grounded in domain-informed assumptions and analytically tractable structure, demonstrates the feasibility of deriving physiologically meaningful insights from limited observational data. Its clear parameter formulation offers transparency and potential for exercise physiological interpretability, which make it a useful foundation for applications where model simplicity, interpretability, and low computational cost are essential. Despite these advantages, the PM framework has limitations. As a model with only one state, it is inherently restricted in its ability to capture dynamic or time-varying physiological processes and to cover the variability that is seen in the data, which are often critical for understanding real-world behaviors such as adaptation, fatigue, or recovery. Additionally, the model's reliance on pre-defined functional forms and bounded parameter spaces can lead to convergence towards edge solutions, as observed in several participants. Future work could enhance the model by extending it into a time-aware or state-space formulation to capture dynamic cardiovascular transitions. Incorporating hierarchical or Bayesian frameworks may also help quantify uncertainty in parameter estimation while preserving interpretability.

In the PM, parameters $\lambda_1$-$\lambda_6$ carries distinct physiological implications. $\lambda_1$ and $\lambda_2$ govern the logarithmic relationship between SV and $\dot{V}O_2$, where $\lambda_1$ may reflect the dynamic enhancement of myocardial contractility during exercise, while $\lambda_2$ could represent baseline SV at rest. Similarly, $\lambda_3$ and $\lambda_4$, associated with TPR and $\dot{V}O_2$, likely characterize vascular adaptability to metabolic demands. $\lambda_3$ indicates the rate of TPR reduction with increasing exercise intensity, and $\lambda_4$ reflects baseline vascular tone. $\lambda_5$ and $\lambda_6$ link HR and MAP changes, may map ANS regulation efficiency, with $\lambda_5$ representing sympathetic responsiveness to blood pressure fluctuations and $\lambda_6$ potentially denoting parasympathetic baseline inhibition. The individualized variability in these parameters (e.g., broader $\lambda$ ranges in PMB-NN vs. PM in Appendix E) quantifies distinct physiological regulatory traits. For example, healthy young adults might exhibit higher $\lambda_1$ and $\lambda_6$ which indicates robust cardiovascular compensation, whereas elderly individuals may show reduced $|\lambda_3|$ (impaired vascular resistance modulation), reduced $\lambda_2$ (lower resting SV) and elevated $\lambda_4$ (higher vascular resistance). In cardiovascular disease populations, the HR–$\dot{V}O_2$ coupling can deviate from near-linearity. Abnormally low $\lambda_5$ could reflect autonomic dysfunction, while deviations in $\lambda_3/\lambda_4$ might reflect endothelial impairment. PMB-NN is expected to address this by learning subject specific gains, offsets, and delays from a short calibration bout, enabling piecewise mappings under physiological priors. Integrating these parameters with interpretable AI frameworks is potential for enhancing personalized predictions and enabling mechanistic insights for clinical applications.

### 4.3. Limitations and future work

While this study establishes the theoretical and computational validity of the PMB-NN framework, several limitations inherent to this intermediate stage of development must be addressed to facilitate its translation into clinical practice.

#### 4.3.1. Physiological assumptions and pathological generalization

For the simplified physiological model, the assumptions of SV/TPR-$\dot{V}O_2$ and $\Delta$HR-$\Delta$MAP relationships were made based on data observations from previous studies involving normal subjects. Although we expect these mathematical relationships to be applicable across all populations, where different patient groups are characterized by varying parameter values ($\lambda_1$ to $\lambda_6$), there are situations where these assumptions may not hold. Specifically, in certain patients with more severe conditions, the expected relationships may not be applicable. For example, Parkinson's disease (PD) patients typically exhibit a diminished cardiovascular response during exercise, with a blunted increase in heart rate and blood pressure [41]. Similarly, patients with multiple system atrophy (MSA) often present with autonomic dysfunction, leading to abnormal circadian rhythms in heart rate and blood pressure [42]. Patients with complete atrioventricular block (AVB) may experience arrhythmias during exercise [43]. In these cases, the $\Delta$HR-$\Delta$MAP relationship may not be applicable. Furthermore, in heart failure with preserved ejection fraction (HFpEF), the increase in stroke volume during exercise does not adequately match the metabolic demand [44], suggesting that the SV-$\dot{V}O_2$ relationship may not apply in this group. Uncontrolled hypertension patients often exhibit abnormal peripheral resistance and cardiovascular responses during exercise [45], which may render the TPR-$\dot{V}O_2$ relationship less applicable. On the other hand, the current sole PM and PMB-NN were validated within the submaximal range, which have not been validated on datasets from extreme exercise scenarios, or patients with cardiovascular diseases or autonomic dysfunction. Importantly, our proposed PMB-NN framework does not assume that the model remains valid under pathological conditions. Instead, the model serves as a personalized reference of expected behavior under nominal physiology, against which observed HR trajectories can be compared. The role of the digital twin is therefore not to diagnose pathology directly, but to provide a stable baseline that enables deviations to be detected relative to an individual-specific norm. Future work could focus on developing disease-specific adjustments to the mathematical model to account for variations in cardiovascular responses in patients. Collecting data from high-intensity interval training or maximal exertion tests scenarios could validate model's performance under more demanding conditions and further assess its ability to predict cardiovascular responses across a wider range of exercise intensities. Collaborating with hospital partners to collect continuous oxygen uptake and heart rate data from patient groups during daily life scenarios will enable external validation of the proposed model in clinical populations and allow us to assess its potential as an early-warning tool for diseases.

#### 4.3.2. Generalization of latent parameters

In this study, each personalized model estimates individual-specific latent parameters that capture subject-dependent physiological characteristics. These latent representations, however, are not directly generalizable across individuals. In future work, we aim to extend this framework by collecting data from a larger number of participants (e.g., hundreds) to analyze the structure of these individualized latent spaces at the population level. By aggregating and clustering the personalized latent variables, we expect to uncover shared patterns or subtypes that reflect underlying physiological or behavioral differences. Building on these insights, we plan to develop a hierarchical or mixture-of-experts model that learns a population prior over the individual latent parameters, allowing new users to be initialized from these learned priors and then rapidly fine-tuned using a short personal calibration clip. Such an approach would preserve the interpretability and adaptability of personalized models while enhancing their scalability and generalization for broader clinical deployment.

#### 4.3.3. Transition from measured to estimated metabolic input

A primary limitation of the current study design, intended as an intermediate validation step, is the reliance on direct metabolic measurement ($\dot{V}O_2$) as the model input. While measuring $\dot{V}O_2$ offers the high precision necessary to validate the PMB-NN's core efficacy, it is impractical for daily life monitoring. To address this, our future work will integrate the present framework with our team's ongoing developing accelerometer-based energy expenditure estimation models [46], creating a seamless IMU-to-HR pipeline capable for daily life monitoring.

#### 4.3.4. Model optimization and architecture

Our current model faces challenges in balancing the data loss and physiological loss terms. Increasing the weight of the data loss improves the fitting accuracy to observed HR values but may compromise the precision of parameter identification. Conversely, emphasizing the physiological loss enhances parameter interpretability and adherence to physiological constraints but can reduce HR estimation accuracy. To resolve the conflict between mathematical stability and physiological realism caused by loss function imbalance, a dynamic weighting scheme could be implemented to prioritize equation constraints ($L_{DE}$) during stable physiological states while emphasizing data fidelity ($L_{data}$) during dynamic activities, thereby mitigating parameter clustering artifacts. Moreover, architectural improvements such as replacing Tanh activations with Leaky Rectified Linear Unit (Leaky ReLU), could alleviate gradient saturation in autonomic terms ($\lambda_5$-$\lambda_6$) and enhance the model's ability to capture abrupt physiological transitions, such as HR spikes during 'sit-to-stand' maneuvers in WEEE-02 and 03. Additionally, embedding hybrid mechanistic-data layers, such as incorporating the Windkessel model for TPR dynamics as hard constraints, could improve sensitivity to weak monotonicity and stochastic fluctuations then address the oversimplification inherent in the current algebraic–differential framework. Furthermore, leveraging transfer learning from larger cardiovascular datasets could enhance generalization for atypical cases (e.g., WEEE-13's vagal response) while retaining personalized adaptation through fine-tuning on individual exercise patterns.

*4.3.5. External validation via multimodal sensing*

Finally, to address data scarcity and limited external validation, future work will integrate additional non-invasive sensing modalities. In follow-up studies, we plan to collect self-acquired datasets incorporating continuous blood pressure monitoring, which can provide beat-to-beat estimates of SV and TPR (combined with established methods such as Modelflow [47]). These derived measurements will serve as an independent reference for assessing the plausibility and robustness of parameter estimates from the PM, offering a more comprehensive framework for evaluating model validity under varied physiological conditions.

Collectively, these refinements aim to harmonize data-driven learning with physiological coherence, enabling robust, individualized HR estimation across diverse populations and advancing the framework's potential for real-time cardiovascular risk monitoring.

## 5. Conclusion

In this research, we proposed a physiological model-based neural network (PMB-NN) for modeling metabolic–heart rate relationship across various physical activities, serving as an intermediate foundational model for wearable-based daily life monitoring. The PMB-NN demonstrates promising estimation accuracy equivalent to sole data-driven model and outperforms sole physiological model, due to its integration of physiological principles, which enhances predictive accuracy and model interpretability. Meanwhile, the framework maintains robustness against $\dot{V}O_2$ uncertainty within specific operational boundaries (up to 20% noise). Moreover, the simplified physiological model we propose, along with the individualized parameters identified by PMB-NN, effectively reproduced the dynamics of heart rate responses. This study highlights the potential of combining data-driven methods with physiological insights for personalized health monitoring. Future work may involve refining the physiological model and combining advanced network architectures to build a digital twin for early cardiovascular risk detection in home settings.

## CRediT authorship contribution statement

**Yaowen Zhang:** Writing – review & editing, Writing – original draft, Visualization, Validation, Software, Methodology, Formal analysis, Conceptualization. **Libera Fresiello:** Writing – review & editing, Supervision, Methodology. **Peter H. Veltink:** Writing – review & editing, Supervision. **Dirk W. Donker:** Writing – review & editing, Supervision. **Ying Wang:** Writing – review & editing, Supervision, Methodology, Conceptualization.

## Statements of ethical approval

This computational study used only anonymized public datasets, thus requiring no ethics approval per institutional policies.

Citation to the public datasets:

Shkurta Gashi, Chulhong Min, Alessandro Montanari, Silvia Santini, & Fahim Kawsar. . WEEE, A Multi-Device and Multi-Modal Dataset for Wearable Human Energy Expenditure Estimation [Data set]. Zenodo. https://doi.org/10.5281/zenodo.6420886

Mongin, D., García Romero, J., & Alvero Cruz, J. R. . Treadmill Maximal Exercise Tests from the Exercise Physiology and Human Performance Lab of the University of Malaga (version 1.0.0). PhysioNet. RRID:SCR_007345. https://doi.org/10.13026/de8x-qg95

## Declaration of Generative AI and AI-assisted technologies in the writing process

During the preparation of this work, the author(s) used Gemini in order to improve the language clarity, refine the logical flow, and perform spelling and grammar correction. After using this tool/service, the author(s) reviewed and edited the content as needed and take(s) full responsibility for the content of the published article.

## Funding

The current research was partly funded by the China Scholarship Council (CSC).

## Declaration of competing interest

The authors declare the following financial interests/personal relationships which may be considered as potential competing interests: Yaowen Zhang reports financial support was provided by China Scholarship Council (CSC). If there are other authors, they declare that they have no known competing financial interests or personal relationships that could have appeared to influence the work reported in this paper.

## Appendix A. Data preprocessing consequence

An example of data preprocessing consequence is shown in Fig. 8.

## Appendix B. Validation of curve fitting for stroke volume and total peripheral resistance

In Fig. 9(a), we presented the fitted logarithmic curves to the data about the SV and TPR response to $\dot{V}O_2$ in [36], whose SV and TPR data were derived from hemodynamic analysis of HR and ABP recordings. The data had displayed SV and TPR changes within different intensities from resting to submaximal exercise (unit: kpm/min) for 12 healthy participants. We converted the unit from exercise intensity to oxygen uptake according to the formula described in [48]. The $R^2$ of fit for SV and TPR are 0.98 and 0.99, respectively. Therefore, initial values of 0.02, 0.1, −5.3, and 10.5 were chosen for $\lambda_1$ through $\lambda_4$, respectively.

Normal SV values range is 0.057 to 0.144 L/min [49], while TPR typically falls between 8.75 to 20 mmHg/L/min [36,50,51]. It is expected that the estimated SV and TPR will lie within these physiologically explainable intervals. Accordingly, we examined the feasible ranges for the unknown parameters $\lambda_1$ to $\lambda_4$ within the proposed logarithmic functions, guided by the normal ranges of SV, TPR, and $\dot{V}O_2$ reported in [36]. The approximate intervals are: $\lambda_1$ (0.01, 0.03), $\lambda_2$ (0.06, 0.15), $\lambda_3$ (−6, −2), and $\lambda_4$ (7, 20). In Fig. 9(b), the explainable ranges of the SV-$\dot{V}O_2$ and TPR-$\dot{V}O_2$ relationships are illustrated.

## Appendix C. Validation of linear relationship between the change of HR and change of MAP

Verification of the linear relationship between $\Delta$HR and $\Delta$MAP was done based on the data in [36]. A continuous 3 minutes' time pattern of HR and MAP wave was simulated with their model in transient phase from rest to moderate exercise. We applied their data to express HR and MAP waveform (MAP = 1/3 SBP + 2/3 DBP) shown in Fig. 10(a), resampled at 1 Hz on each waveform (0–180 s, 181 points), calculated $\Delta$HR and $\Delta$MAP were obtained by differentiation from HR and MAP and showed in the hexagonal binning plot of $\Delta$HR and $\Delta$MAP in Fig. 10(b). We disregarded the rapid drop in MAP between 65 and 70 s, which was caused by the time delay in the decrease of TPR following the increase in exercise intensity at 1 min, owing to the omission of time delay dynamics in the simplified TPR-$\dot{V}O_2$ model in Eq. (4).

In Fig. 10(b), the linear fit effectively captured the relationship between $\Delta$HR and $\Delta$MAP. A total of 167 out of 180 data points (92.78%) fell within the 95% confidence interval, indicating that the majority of the points are closely aligned with the linear regression model. In Fig. 10(c), the explainable ranges of $\lambda_5$ and $\lambda_6$ in $\Delta$HR-$\Delta$MAP relationship is illustrated. HR changes are generally positively correlated with MAP variations, especially within the regulation mediated by the baroreflex [52]. Based on this prior empirical observation in exercise HR and

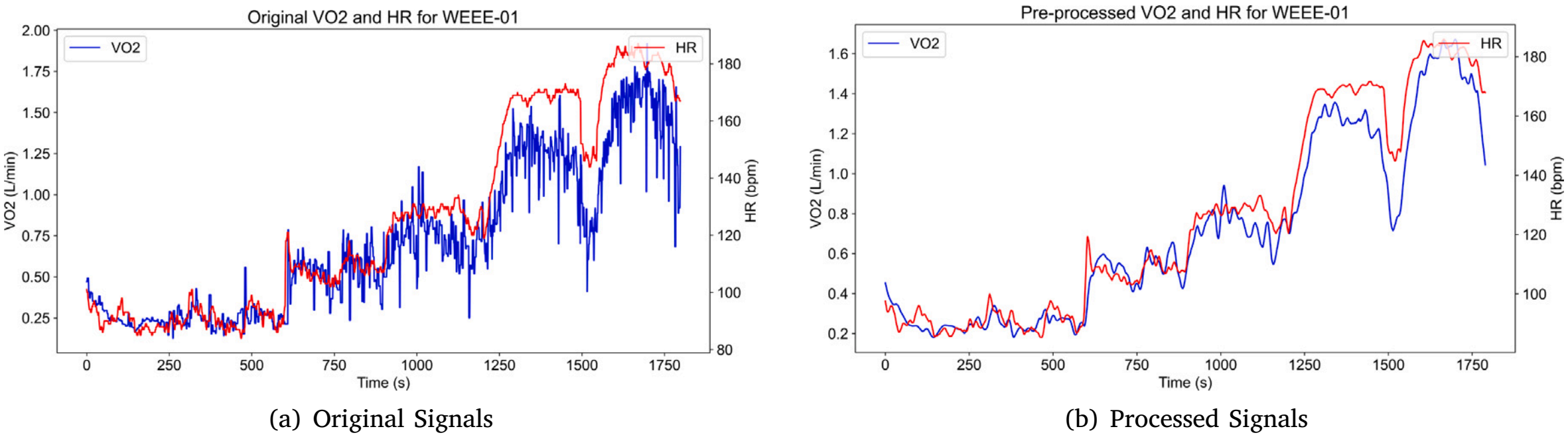

(a) Original Signals

(b) Processed Signals

**Fig. 8.** The measured signals and filtered signals of one subject. $\dot{V}O_2$ (blue) and HR (red) signals from measurement 8(a) and preprocessed 8(b) are plotted in the figure.

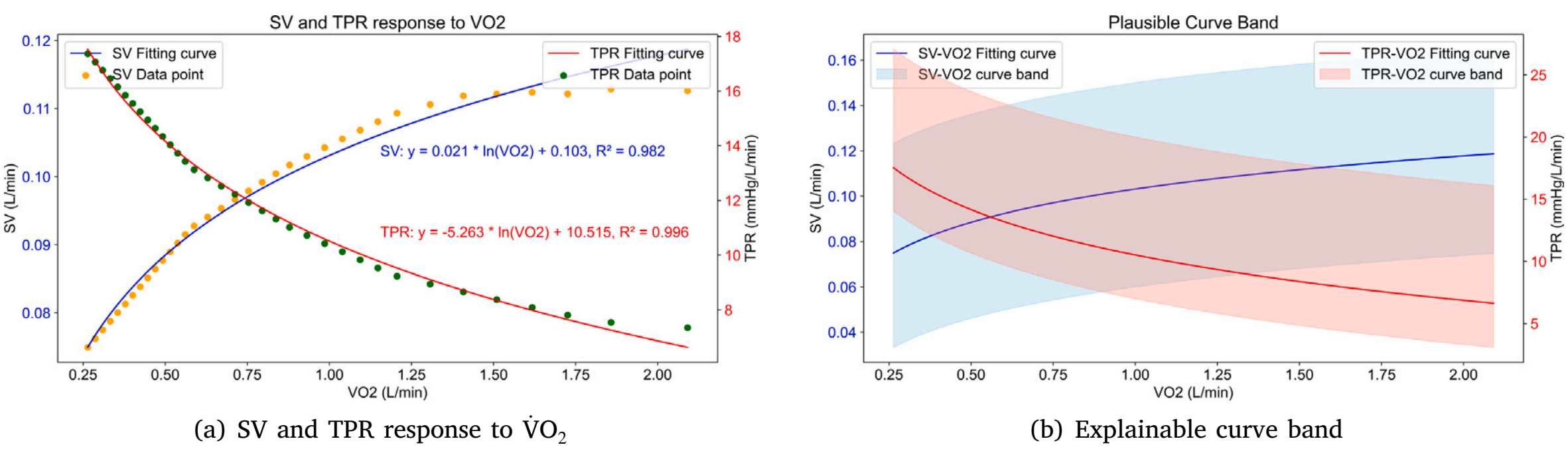

(a) SV and TPR response to $\dot{V}O_2$

(b) Explainable curve band

**Fig. 9.** 9(a): Logarithmic Regression Curves and Functions for SV and TPR. Orange and green points are SV and TPR data replicated from [36], respectively. Blue and red lines indicate our fitted curves that represent SV and TPR response to $\dot{V}O_2$, respectively. 9(b): Possible bands of SV-$\dot{V}O_2$ (in skyblue) and TPR-$\dot{V}O_2$ (in salmon) relationship curve, with ranges as: $\lambda_1$ (0.01, 0.03), $\lambda_2$ (0.06, 0.15), $\lambda_3$ (−6, -2), $\lambda_4$ (7, 20) from Eq. (6).

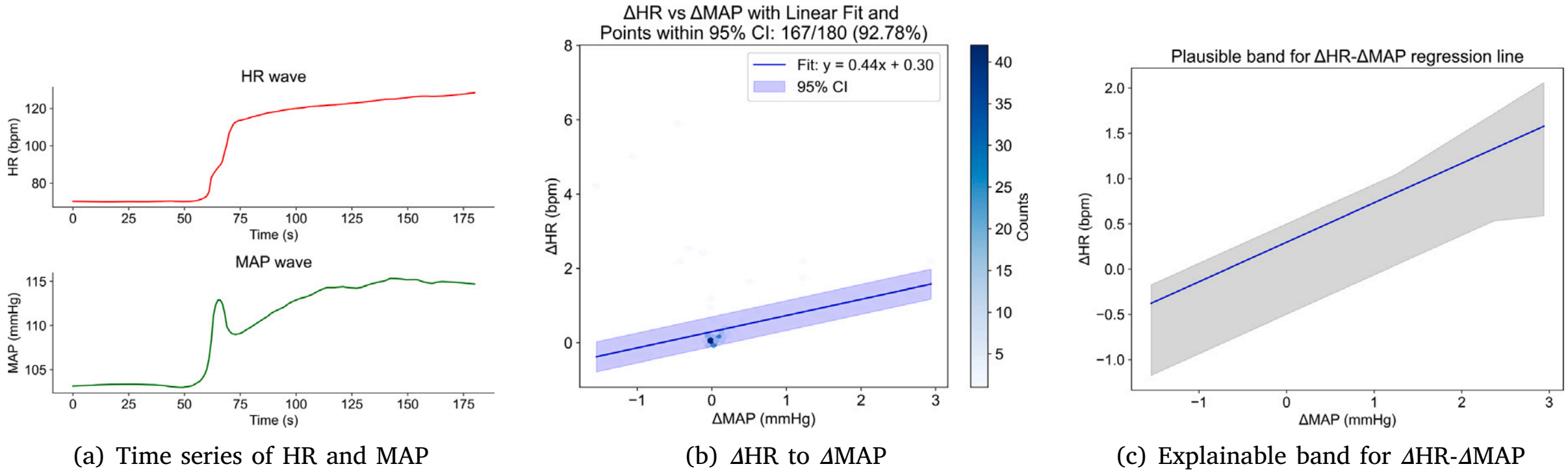

(a) Time series of HR and MAP

(b) $\Delta$HR to $\Delta$MAP

(c) Explainable band for $\Delta$HR-$\Delta$MAP

**Fig. 10.** Validation of linear relation between $\Delta$HR and $\Delta$MAP. 10(a): Time series of heart rate in red and mean arterial pressure in green. 10(b): Linear regression result of $\Delta$HR and $\Delta$MAP. Blue hexagon points with different color depth represent amount and density of $\Delta$MAP and $\Delta$HR. Blue straight line is the linear regression curve for $\Delta$HR and $\Delta$MAP and light blue area is the 95% confidence interval. 10(c): Explainable range of the $\Delta$HR-$\Delta$MAP regression line. The blue line represents the current fit, and the shaded area indicates the range of possible lines within the specified parameter bounds. The parameter ranges are (0.1, 0.6) for $\lambda_5$ and (−0.5, 0.5) for $\lambda_6$.

**Table 2**
Performance metrics for PMB-NN, FCNN, and PM models across participants.

| Participant | PMB-NN | | FCNN | | PM | |
|---|---|---|---|---|---|---|
| | $R^2$ | RMSE | $R^2$ | RMSE | $R^2$ | RMSE |
| WEEE-01 | 0.98 | 4.88 | 0.98 | 4.44 | 0.95 | 8.03 |
| WEEE-02 | 0.70 | 13.58 | 0.70 | 13.49 | 0.07 | 23.90 |
| WEEE-03 | 0.69 | 9.34 | 0.59 | 10.61 | 0.28 | 14.12 |
| WEEE-04 | 0.90 | 9.66 | 0.91 | 9.43 | 0.73 | 16.67 |
| WEEE-05 | 0.94 | 6.90 | 0.97 | 5.07 | 0.18 | 24.79 |
| WEEE-06 | 0.73 | 7.23 | 0.69 | 7.77 | 0.69 | 7.80 |
| WEEE-08 | 0.93 | 6.24 | 0.95 | 5.97 | 0.59 | 17.01 |
| WEEE-09 | 0.70 | 18.08 | 0.74 | 16.78 | 0.55 | 22.13 |
| WEEE-12 | 0.97 | 5.15 | 0.98 | 3.88 | 0.91 | 8.61 |
| WEEE-13 | −0.65 | 14.30 | −0.49 | 13.09 | −1.74 | 17.78 |
| WEEE-15 | 0.73 | 14.91 | 0.76 | 13.97 | 0.77 | 13.83 |
| WEEE-17 | 0.95 | 6.84 | 0.96 | 6.48 | −0.35 | 36.30 |
| ACTES-1 | 0.79 | 16.37 | 0.76 | 17.30 | 0.38 | 28.04 |
| ACTES-2 | 0.95 | 8.52 | 0.99 | 4.38 | 0.79 | 17.53 |
| ACTES-3 | 0.93 | 7.06 | 0.95 | 6.09 | −0.44 | 33.12 |
| ACTES-4 | 0.85 | 10.57 | 0.92 | 7.58 | −0.25 | 30.77 |
| ACTES-5 | 0.89 | 9.96 | 0.96 | 5.96 | 0.88 | 10.32 |
| ACTES-7 | 0.88 | 10.71 | 0.90 | 9.69 | 0.58 | 19.63 |
| ACTES-9 | 0.80 | 13.69 | 0.81 | 13.15 | 0.57 | 19.84 |
| ACTES-10 | 0.91 | 10.43 | 0.85 | 13.42 | 0.93 | 9.05 |
| ACTES-12 | 0.95 | 6.54 | 0.77 | 13.99 | 0.70 | 16.06 |
| ACTES-13 | 0.84 | 11.54 | 0.88 | 9.91 | −0.02 | 28.74 |
| ACTES-14 | 0.73 | 15.40 | 0.71 | 15.92 | 0.77 | 14.18 |
| ACTES-15 | 0.85 | 13.17 | 0.86 | 12.97 | 0.86 | 12.87 |
| ACTES-18 | 0.90 | 11.77 | 0.99 | 3.88 | 0.54 | 24.64 |
| median | 0.88 | 9.96 | 0.88 | 9.69 | 0.58 | 17.53 |
| max | 0.99 | 18.08 | 0.99 | 17.30 | 0.95 | 36.30 |
| min | −0.65 | 4.54 | −0.49 | 3.88 | −1.74 | 7.80 |
| p value | – | – | 0.305 | 0.059 | 0.000 | 0.000 |
| d value | – | – | 0.018 | −0.075 | −0.832 | 1.398 |

MAP dynamics accompanied with the fitted values 0.44 and 0.3, we defined the positive range (0.1, 0.6) for $\lambda_5$, while (−0.5, 0.5) for $\lambda_6$ which distributed around zero in both positive and negative directions.

## Appendix D. Individualized performance from different models and comparison

Table 2 listed the HR estimation accuracy for each participant from three models in $R^2$, RMSE. Table 3 listed the Spearman's rank correlation coefficient ($\rho$) for each participant from three models. Median, max, min values for each metric and the statistical comparison between PMB-NN and FCNN, between PMB-NN and PM were also listed in both Tables.

## Appendix E. Personally identified values of unknown parameters in mathematical model from both PMB-NN and PM

Table 4 listed the identified $\lambda$ values for each participant form PMB-NN and the reconstructed HR accuracy with the $\lambda$s (PMB-NN-R), while Table 5 listed the identified $\lambda$ values for each participant form PM.

## Appendix F. Statistical summary and comparative analysis of estimation results across three models in different exercise conditions

The detailed activity-segmented $R^2$ and RMSE values for all models across WEEE participants, and the one-tailed p-values for pairwise comparisons within each activity type between PMB-NN and FCNN, as well as PMB-NN and PM are provided in Table 6. P-values and Cohen's d-values were derived from pairwise comparisons between PMB-NN and the respective models for identical metrics under each exercise condition.

## Appendix G. Estimation results on participant WEEE-06 from PMB-NN-R and PM

See Fig. 11.

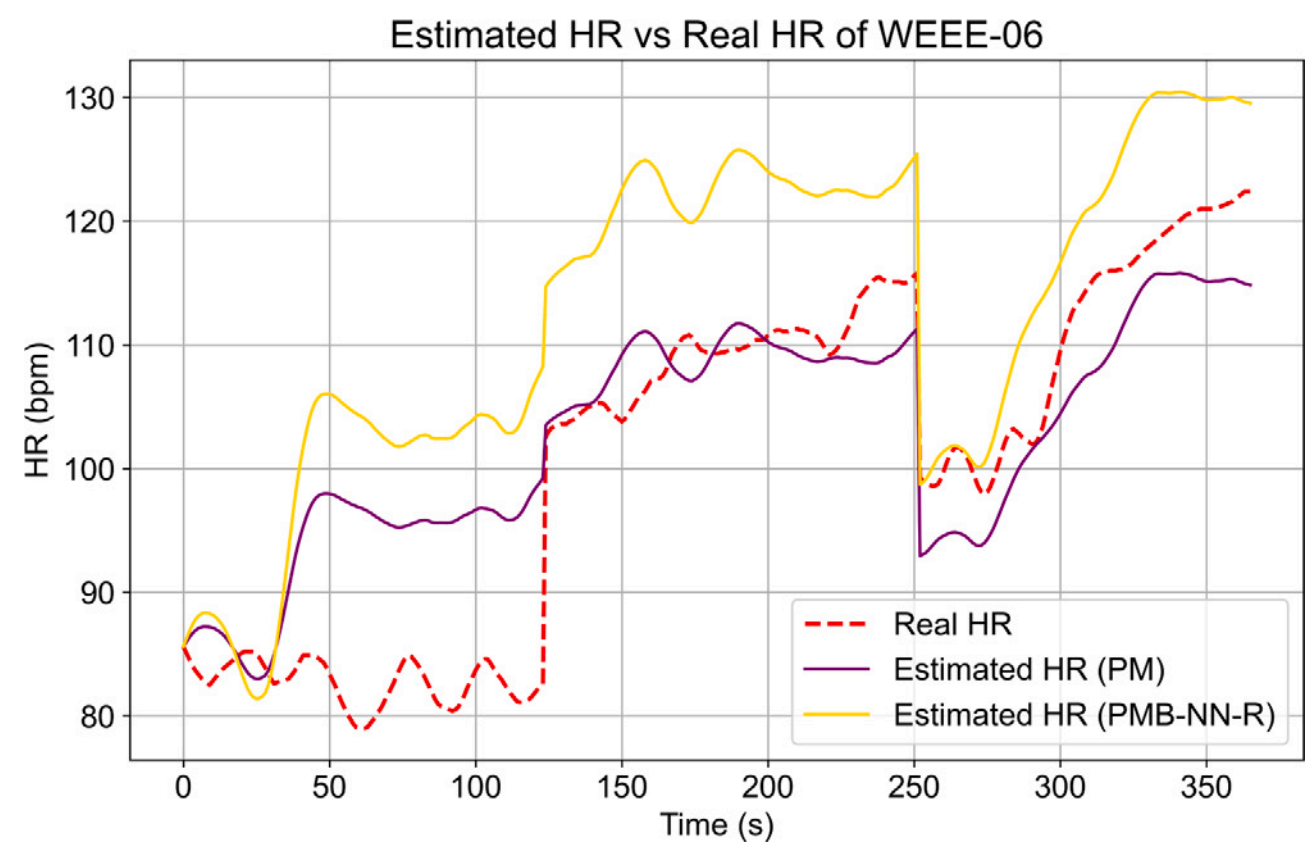

**Fig. 11.** PMB-NN-R's and PM's HR estimation results of WEEE-06. The red dash curve in each plot represents the measured HR values, while the purple and gold solid curves represent the PM and PMB-NN-R estimated HR values, respectively.

## Appendix H. Pre-processed data for participant WEEE-02

Fig. 12 provided a sample of the same sudden increasing trend from sitting to standing among 8 participants' data.

## Appendix I. Summary of HR estimation studies

The method, dataset, input(s) variables and accuracy in previous HR estimation studies and our current study are summarized in .

**Table 3**
Spearman's rank correlation coefficient ($\rho$) between $\dot{V}O_2$ and estimated HR for three models.

| Participant | PMB-NN | FCNN | PM |
|---|---|---|---|
| WEEE-01 | 1 | 1 | 0.976 |
| WEEE-02 | 1 | 1 | 0.996 |
| WEEE-03 | 0.999 | 1 | 0.996 |
| WEEE-04 | 1 | 1 | 0.947 |
| WEEE-05 | 1 | 0.996 | 0.993 |
| WEEE-06 | 1 | 1 | 0.999 |
| WEEE-08 | 1 | 1 | 0.965 |
| WEEE-09 | 1 | 1 | 0.935 |
| WEEE-12 | 1 | 1 | 0.999 |
| WEEE-13 | 1 | 1 | 0.989 |
| WEEE-15 | 1 | 0.986 | 0.963 |
| WEEE-17 | 1 | 1 | 0.996 |
| ACTES-1 | 1 | 1 | 0.996 |
| ACTES-2 | 1 | 1 | 0.996 |
| ACTES-3 | 1 | 0.866 | 0.948 |
| ACTES-4 | 1 | 1 | 1 |
| ACTES-5 | 1 | 1 | 1 |
| ACTES-7 | 1 | 1 | 0.979 |
| ACTES-9 | 1 | 1 | 0.999 |
| ACTES-10 | 1 | 0.953 | 1 |
| ACTES-12 | 1 | 1 | 1 |
| ACTES-13 | 1 | 0.982 | 1 |
| ACTES-14 | 1 | 0.863 | −0.162 |
| ACTES-15 | 1 | 1 | 0.923 |
| ACTES-18 | 1 | 1 | 0.977 |
| median | 1 | 1 | 0.996 |
| max | 1 | 1 | 1 |
| min | 0.999 | 0.863 | −0.162 |
| *p*-value | – | 0.028 | 0 |
| d-value | – | −0.372 | −0.276 |

**Table 4**
PMB-NN identified lambda values and PMB-NN-R's performance.

| Participant | PMB-NN | | | | | | PMB-NN-R | |
|---|---|---|---|---|---|---|---|---|
| | $\lambda_1$ | $\lambda_2$ | $\lambda_3$ | $\lambda_4$ | $\lambda_5$ | $\lambda_6$ | $R^2$ | RMSE |
| WEEE-01 | 0.017 | 0.109 | −2.000 | 19.921 | 0.404 | 0.000 | 0.90 | 11.62 |
| WEEE-02 | 0.012 | 0.144 | −2.000 | 19.720 | 0.366 | −0.001 | 0.66 | 14.31 |
| WEEE-03 | 0.015 | 0.118 | −2.000 | 19.926 | 0.362 | −0.001 | 0.38 | 13.09 |
| WEEE-04 | 0.020 | 0.109 | −2.000 | 19.553 | 0.385 | 0.002 | 0.80 | 14.50 |
| WEEE-05 | 0.016 | 0.125 | −2.000 | 19.786 | 0.357 | 0.002 | 0.70 | 14.92 |
| WEEE-06 | 0.016 | 0.129 | −2.000 | 19.675 | 0.349 | 0.000 | 0.04 | 13.73 |
| WEEE-08 | 0.016 | 0.141 | −2.001 | 19.585 | 0.327 | 0.000 | 0.54 | 18.11 |
| WEEE-09 | 0.020 | 0.110 | −2.000 | 19.834 | 0.351 | 0.002 | 0.45 | 24.55 |
| WEEE-12 | 0.020 | 0.122 | −2.005 | 18.477 | 0.382 | −0.001 | 0.93 | 7.56 |
| WEEE-13 | 0.014 | 0.075 | −2.005 | 19.498 | 0.142 | 0.000 | −0.27 | 12.11 |
| WEEE-15 | 0.021 | 0.105 | −2.000 | 19.901 | 0.332 | 0.000 | 0.69 | 15.93 |
| WEEE-17 | 0.023 | 0.135 | −2.000 | 19.663 | 0.303 | 0.002 | 0.14 | 29.02 |
| ACTES-1 | 0.021 | 0.126 | −2.427 | 18.634 | 0.394 | 0.002 | 0.68 | 20.15 |
| ACTES-2 | 0.013 | 0.119 | −2.000 | 19.847 | 0.414 | 0.002 | 0.65 | 22.86 |
| ACTES-3 | 0.023 | 0.124 | −2.033 | 19.402 | 0.347 | 0.001 | 0.31 | 22.96 |
| ACTES-4 | 0.025 | 0.135 | −2.094 | 19.126 | 0.327 | −0.001 | 0.83 | 11.37 |
| ACTES-5 | 0.023 | 0.124 | −2.031 | 19.356 | 0.353 | −0.001 | 0.87 | 10.96 |
| ACTES-7 | 0.023 | 0.115 | −2.842 | 17.886 | 0.445 | 0.001 | 0.80 | 13.70 |
| ACTES-9 | 0.024 | 0.136 | −2.014 | 18.675 | 0.320 | 0.001 | −0.34 | 35.23 |
| ACTES-10 | 0.021 | 0.136 | −2.000 | 17.708 | 0.380 | −0.001 | 0.15 | 31.50 |
| ACTES-12 | 0.022 | 0.110 | −2.000 | 19.706 | 0.369 | 0.001 | 0.71 | 15.80 |
| ACTES-13 | 0.024 | 0.112 | −2.017 | 19.487 | 0.357 | 0.001 | 0.78 | 13.22 |
| ACTES-14 | 0.021 | 0.128 | −2.098 | 18.892 | 0.371 | −0.001 | 0.70 | 16.39 |
| ACTES-15 | 0.014 | 0.126 | −2.000 | 19.898 | 0.388 | −0.003 | 0.16 | 31.71 |
| ACTES-18 | 0.015 | 0.141 | −2.000 | 19.364 | 0.364 | −0.001 | 0.47 | 26.45 |

**Table 5**
PM identified lambda values.

| Participant | PM | | | | | |
|---|---|---|---|---|---|---|
| | $\lambda_1$ | $\lambda_2$ | $\lambda_3$ | $\lambda_4$ | $\lambda_5$ | $\lambda_6$ |
| WEEE-01 | 0.030 | 0.060 | −2.000 | 20.000 | 0.370 | 0.014 |
| WEEE-02 | 0.030 | 0.060 | −2.000 | 20.000 | 0.100 | 0.024 |
| WEEE-03 | 0.014 | 0.060 | −2.000 | 19.598 | 0.571 | 0.000 |
| WEEE-04 | 0.030 | 0.060 | −2.000 | 20.000 | 0.341 | 0.000 |
| WEEE-05 | 0.030 | 0.060 | −3.135 | 8.475 | 0.463 | 0.023 |
| WEEE-06 | 0.030 | 0.060 | −2.000 | 20.000 | 0.258 | −0.003 |
| WEEE-08 | 0.030 | 0.060 | −2.000 | 20.000 | 0.346 | −0.003 |
| WEEE-09 | 0.030 | 0.060 | −2.000 | 20.000 | 0.346 | 0.001 |
| WEEE-12 | 0.030 | 0.060 | −2.000 | 20.000 | 0.378 | −0.002 |
| WEEE-13 | 0.010 | 0.150 | −3.591 | 7.000 | 0.100 | 0.023 |
| WEEE-15 | 0.030 | 0.060 | −2.000 | 20.000 | 0.224 | 0.024 |
| WEEE-17 | 0.030 | 0.060 | −2.000 | 20.000 | 0.100 | 0.019 |
| ACTES-1 | 0.030 | 0.070 | −3.625 | 14.489 | 0.600 | 0.039 |
| ACTES-2 | 0.030 | 0.060 | −3.559 | 18.981 | 0.495 | 0.039 |
| ACTES-3 | 0.030 | 0.063 | −3.488 | 16.624 | 0.600 | 0.014 |
| ACTES-4 | 0.030 | 0.060 | −2.000 | 20.000 | 0.268 | 0.026 |
| ACTES-5 | 0.030 | 0.085 | −3.003 | 16.923 | 0.502 | 0.010 |
| ACTES-7 | 0.030 | 0.060 | −2.000 | 20.000 | 0.434 | 0.020 |
| ACTES-9 | 0.030 | 0.062 | −3.940 | 15.878 | 0.586 | 0.039 |
| ACTES-10 | 0.030 | 0.060 | −2.091 | 20.000 | 0.392 | 0.026 |
| ACTES-12 | 0.030 | 0.060 | −2.000 | 20.000 | 0.324 | 0.012 |
| ACTES-13 | 0.030 | 0.060 | −2.000 | 20.000 | 0.217 | 0.023 |
| ACTES-14 | 0.030 | 0.060 | −2.000 | 20.000 | 0.352 | 0.034 |
| ACTES-15 | 0.030 | 0.076 | −2.774 | 17.171 | 0.464 | 0.030 |
| ACTES-18 | 0.030 | 0.060 | −3.640 | 17.057 | 0.539 | 0.030 |

**Table 6**
Estimation and comparison results summary for different activities.

| Participant | Resting | | | | | | Cycling | | | | | | Running | | | | | |
|---|---|---|---|---|---|---|---|---|---|---|---|---|---|---|---|---|---|---|
| | PMB-NN | | FCNN | | PM | | PMB-NN | | FCNN | | PM | | PMB-NN | | FCNN | | PM | |
| | $R^2$ | RMSE | $R^2$ | RMSE | $R^2$ | RMSE | $R^2$ | RMSE | $R^2$ | RMSE | $R^2$ | RMSE | $R^2$ | RMSE | $R^2$ | RMSE | $R^2$ | RMSE |
| WEEE-01 | −0.75 | 5.02 | 0.02 | 3.76 | −1.05 | 5.44 | −8.46 | 5.24 | −13.03 | 6.38 | −4.68 | 4.06 | 0.17 | 4.40 | 0.79 | 2.23 | −4.93 | 11.79 |
| WEEE-02 | −0.98 | 4.11 | −0.88 | 4.00 | −3.27 | 6.03 | −0.45 | 9.39 | −0.70 | 10.20 | −1.09 | 11.29 | −9.46 | 22.26 | −8.96 | 21.72 | −35.17 | 41.40 |
| WEEE-03 | −11.64 | 10.55 | −1.93 | 5.07 | −31.45 | 16.90 | −9.18 | 3.80 | −4.68 | 2.84 | −122.49 | 13.24 | −0.17 | 11.48 | −3.04 | 21.31 | 0.69 | 5.95 |
| WEEE-04 | −0.69 | 9.53 | 0.03 | 7.21 | −1.52 | 11.63 | 0.29 | 5.75 | −0.39 | 8.08 | 0.20 | 6.14 | −68.35 | 17.65 | −35.18 | 12.75 | −162.52 | 27.11 |
| WEEE-05 | −0.96 | 7.82 | 0.61 | 3.50 | 0.24 | 4.89 | 0.76 | 3.66 | 0.84 | 3.02 | −1.42 | 11.70 | −213.33 | 8.77 | −195.36 | 8.39 | −6176.94 | 47.06 |
| WEEE-06 | −32.09 | 9.88 | −7.77 | 5.08 | −47.82 | 12.00 | −1.48 | 5.73 | −5.73 | 9.43 | 0.22 | 3.22 | 0.68 | 4.97 | 0.14 | 8.13 | 0.65 | 5.20 |
| WEEE-08 | −40.20 | 10.51 | −5.32 | 4.05 | −37.42 | 9.98 | 0.68 | 4.09 | 0.91 | 2.11 | −10.68 | 24.69 | −25.69 | 6.21 | −150.86 | 10.06 | −200.90 | 11.60 |
| WEEE-09 | −12.65 | 8.13 | −5.97 | 4.51 | −201.48 | 24.30 | 0.62 | 9.14 | −0.13 | 15.68 | −1.07 | 21.21 | −719.48 | 38.77 | −554.81 | 34.05 | −147.94 | 17.63 |
| WEEE-12 | 0.26 | 7.03 | 0.80 | 3.68 | 0.60 | 5.18 | 0.93 | 3.31 | 0.87 | 4.65 | 0.84 | 5.19 | −6.87 | 4.01 | −3.67 | 3.09 | −89.28 | 13.58 |
| WEEE-13 | −12.38 | 9.64 | −10.43 | 13.10 | −4.71 | 9.26 | −13.83 | 10.59 | −13.21 | 9.59 | −12.18 | 9.23 | −20.67 | 17.60 | −18.22 | 16.78 | −68.30 | 31.87 |
| WEEE-15 | −12.78 | 7.21 | −22.96 | 9.57 | −1.05 | 2.80 | −1.98 | 4.70 | 0.35 | 2.19 | −4.80 | 6.55 | −5.19 | 28.26 | −3.91 | 25.17 | −4.33 | 26.22 |
| WEEE-17 | −2.86 | 4.77 | −5.67 | 6.73 | −4.96 | 6.36 | −12.27 | 8.22 | −16.36 | 7.41 | −427.88 | 36.85 | −5.33 | 5.78 | −2.59 | 4.16 | −644.57 | 55.82 |
| median | −6.31 | 8.83 | −1.40 | 4.28 | −17.36 | 10.80 | −0.08 | 5.48 | −0.55 | 7.23 | −1.26 | 11.49 | −17.57 | 10.13 | −22.07 | 11.40 | −91.56 | 14.71 |
| max | −0.69 | 10.55 | 0.61 | 7.21 | 0.24 | 24.30 | 0.76 | 9.39 | 0.91 | 15.68 | 0.22 | 24.69 | 0.68 | 38.77 | 0.79 | 34.05 | 0.69 | 47.06 |
| min | −40.20 | 4.11 | −7.77 | 3.50 | −201.48 | 4.89 | −13.83 | 3.66 | −16.36 | 2.11 | −427.88 | 3.22 | −719.48 | 4.40 | −554.81 | 2.23 | −6176.94 | 5.20 |
| p value | —— | —— | 0.092 | 0.064 | 0.519 | 0.424 | —— | —— | 0.470 | 0.519 | 0.129 | 0.034 | —— | —— | 0.151 | 0.470 | 0.064 | 0.077 |
| d value | —— | —— | 0.460 | −0.607 | −0.314 | 0.321 | —— | —— | −0.214 | 0.256 | −0.372 | 0.686 | —— | —— | 0.130 | −0.047 | −0.311 | 0.548 |

**Table 7**
Summary of HR estimation studies: methods, datasets, inputs, and reported accuracy.

| Study | Method | Dataset | Input(s) | Accuracy |
|---|---|---|---|---|
| [19] | LSTM–BiLSTM–Attention | 2 subjects, sleeping | $HR(t-n), \ldots,$ $HR(t-1)$ | RMSE = 2.52 bpm |
| [20] | Evolutionary neural network | 1 subject, daily activities | Pred. $HR(t-1)$, chest $ACC(t-1)$ | MAE = 4.38 bpm<br>Var = 12.96 bpm |
| [21] | Feedforward neural network | 1 subject, daily activities | $HR(t-1)$, chest $ACC(t-1)$ | MAE = 3.31 bpm<br>Var = 18.68 bpm |
| [22] | Feedforward neural network | 7 subjects, physical exercises | $HR(t-1)$, cycle cadence$(t-1)$ | MAE = 3.02 bpm |
| [18] | Support Vector Regression | 1 subject, physical exercises | $HR(t-n)$, workload$(t-n)$ | MSE = 5.72 $bpm^2$ |

**Table 7** (*continued*).

| Study | Method | Dataset | Input(s) | Accuracy |
|---|---|---|---|---|
| [16] | Wiener filter + phase encoder | 23 subjects, physical exercises | 2 wrist PPG($t$), wrist ACC($t$) | MAE = 1.97 bpm |
| [17] | Peak identification in SCG & GCG | 5 subjects, static postures | Chest IMU($t$) | LOA: ±2.86 bpm |
| [23] | K-Nearest Neighbors Regression | 14 participants, daily activities | PAI($t$) from chest + upper leg IMU($t$) | $R^2$ = 0.81, MAE = 6 bpm |
| Our current study | PMB-NN | 25 subjects, physical exercises | $\dot{V}O_2(t)$ | $R^2$ = 0.88, RMSE = 9.97 bpm, MAE = 8.87 bpm |

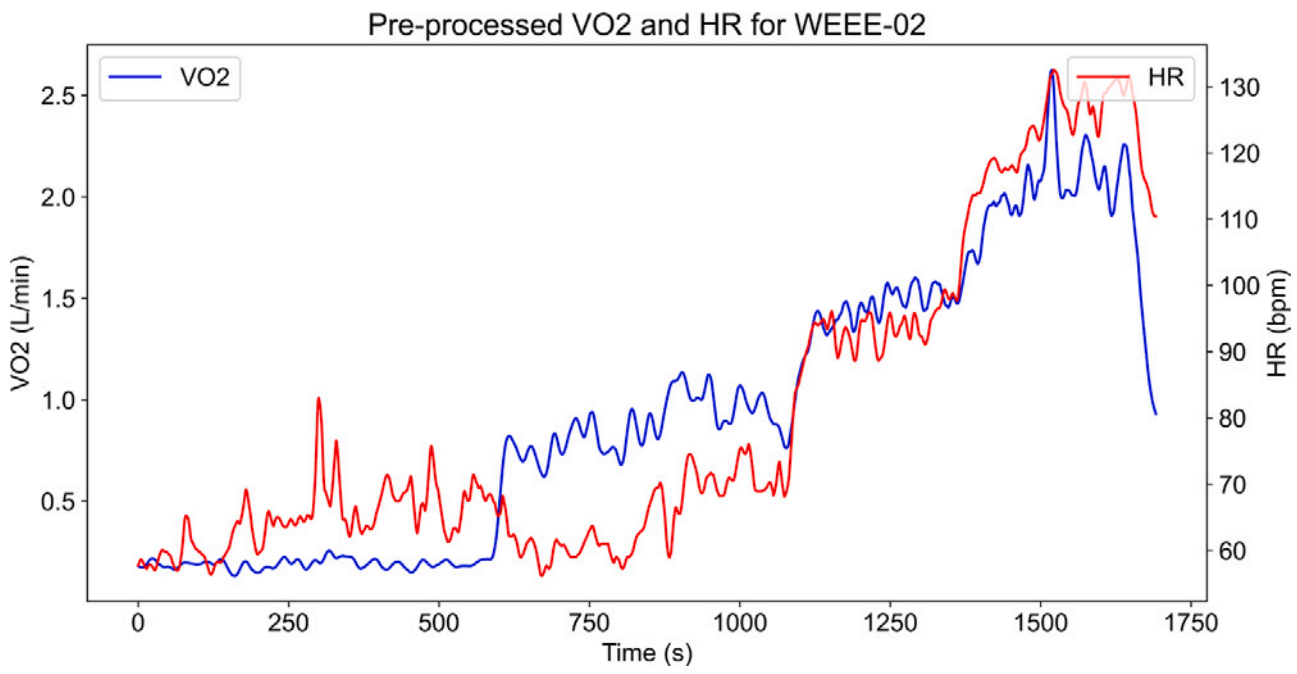

**Fig. 12.** $\dot{V}O_2$ and HR data for participant 2 after preprocessing.

## References

[1] C. Rogers, N. Bush, Heart failure: Pathophysiology, diagnosis, medical treatment guidelines, and nursing management, Nurs. Clin. North Am. 50 (2015) 787–799.

[2] B. Steiner, et al., Challenges in heart failure care in four European countries: a comparative study, Eur. J. Pub. Health 33 (2023) 448–454.

[3] Y. Wang, et al., Early detection of heart failure with varying prediction windows by structured and unstructured data in electronic health records, in: 2015 37th Annual International Conference of the IEEE Engineering in Medicine and Biology Society, EMBC, 2015, pp. 2530–2533.

[4] M. Hori, H. Okamoto, Heart rate as a target of treatment of chronic heart failure, J. Cardiol. 60 (2012) 86–90.

[5] M.T.L. Rovere, J.T. Bigger, F.I. Marcus, A. Mortara, P.J. Schwartz, Baroreflex sensitivity and heart-rate variability in prediction of total cardiac mortality after myocardial infarction, Lancet 351 (1998) 478–484.

[6] D.P. Vivekananthan, E.H. Blackstone, C.E. Pothier, M.S. Lauer, Heart rate recovery after exercise is apredictor of mortality, independent of the angiographic severity of coronary disease, J. Am. Coll. Cardiol. 42 (2003) 831–838.

[7] N.J. Leeper, et al., Prognostic value of heart rate increase at onset of exercise testing, Circulation 115 (2007) 468–474.

[8] C.E. Negrao, H.R. Middlekauff, Adaptations in autonomic function during exercise training in heart failure, Heart Fail. Rev. 13 (2008) 51–60.

[9] P. Lechat, et al., Heart rate and cardiac rhythm relationships with bisoprolol benefit in chronic heart failure in CIBIS II trial, Circulation 103 (2001) 1428–1433.

[10] S.J. Pocock, et al., Predictors of mortality and morbidity in patients with chronic heart failure, Eur. Heart J. 27 (2006) 65–75.

[11] M. Jacobsen, T.A. Dembek, G. Kobbe, P.W. Gaidzik, L. Heinemann, Noninvasive continuous monitoring of vital signs with wearables: Fit for medical use? J. Diabetes Sci. Technol. 15 (2021) 34–43.

[12] G. Quer, P. Gouda, M. Galarnyk, E.J. Topol, S.R. Steinhubl, Inter- and intraindividual variability in daily resting heart rate and its associations with age, sex, sleep, BMI, and time of year: Retrospective, longitudinal cohort study of 92,457 adults, PLoS One 15 (2020) e0227709.

[13] E. Magosso, M. Ursino, Cardiovascular response to dynamic aerobic exercise: A methematical model, Med. Biol. Eng. Comput. 40 (2002) 660–674.

[14] L. Fresiello, B. Meyns, A.D. Molfetta, G. Ferrari, A model of the cardiorespiratory response to aerobic exercise in healthy and heart failure conditions, Front. Physiol. 7 (2016).

[15] C.A. Sarmiento, A.M. Hernández, L.Y. Serna, M.Á. Mañanas, An integrated mathematical model of the cardiovascular and respiratory response to exercise: model-building and comparison with reported models, Am. J. Physiology-Heart Circ. Physiol. 320 (2021) H1235–H1260.

[16] A. Temko, Accurate heart rate monitoring during physical exercises using PPG, IEEE Trans. Biomed. Eng. 64 (2017) 2016–2024.

[17] C. Massaroni, C. Romano, F. De Tommasi, M.B. Čukić, M. Carassiti, D. Formica, E. Schena, Heart rate and heart rate variability indexes estimated by mechanical signals from a skin-interfaced IMU, in: 2022 IEEE International Workshop on Metrology for Industry 4.0 & IoT, MetroInd4.0&IoT, 2022, pp. 322–327.

[18] M. Ludwig, A.M. Sundaram, M. Füller, A. Asteroth, E. Prassler, On modeling the cardiovascular system and predicting the human heart rate under strain, in: Proceedings of the 1st International Conference on Information and Communication Technologies for Ageing Well and e-Health, 2015, pp. 106–117.

[19] H. Lin, S. Zhang, Q. Li, Y. Li, J. Li, Y. Yang, A new method for heart rate prediction based on LSTM-BiLSTM-Att, Measurement 207 (2023) 112384.

[20] F. Xiao, Y. Chen, M. Yuchi, M. Ding, J. Jo, Heart rate prediction model based on physical activities using evolutionary neural network, in: 2010 Fourth International Conference on Genetic and Evolutionary Computing, 2010, pp. 198–201.

[21] M. Yuchi, J. Jo, Heart rate prediction based on physical activity using feedforwad neural network, in: 2008 International Conference on Convergence and Hybrid Information Technology, 2008, pp. 344–350.

[22] K. Mutijarsa, M. Ichwan, D.B. Utami, Heart rate prediction based on cycling cadence using feedforward neural network, in: 2016 International Conference on Computer, Control, Informatics and Its Applications, IC3INA, 2016, pp. 72–76.

[23] M. Thoonen, P.H. Veltink, F.R. Halfwerk, R.W. van Delden, Y. Wang, A movement-artefact-free heart-rate prediction system, in: 2022 Computing in Cardiology Conference, 2022.

[24] G.E. Karniadakis, I.G. Kevrekidis, L. Lu, P. Perdikaris, S. Wang, L. Yang, Physics-informed machine learning, Nat. Rev. Phys. 3 (2021) 422–440.

[25] S. Cai, Z. Wang, S. Wang, P. Perdikaris, G.E. Karniadakis, Physics-informed neural networks for heat transfer problems, J. Heat Transf. 143 (2021).

[26] M. Daneker, Z. Zhang, G.E. Karniadakis, L. Lu, Systems biology: Identifiability analysis and parameter identification via systems-biology-informed neural networks, 2023, pp. 87–105.

[27] S. Gashi, C. Min, A. Montanari, S. Santini, F. Kawsar, A multidevice and multimodal dataset for human energy expenditure estimation using wearable devices, Sci. Data 9 (2022) 537.

[28] D. Mongin, C. Chabert, D.S. Courvoisier, J. García-Romero, J.R. Alvero-Cruz, Heart rate recovery to assess fitness: comparison of different calculation methods in a large cross-sectional study, Res. Sport. Med. 31 (2023) 157–170.

[29] C.R. McNulty, R.A. Robergs, Repeat trial and breath averaging: Recommendations for research of VO2 kinetics of exercise transitions to steady-state, Mov. Sport. Sci. - Sci. Motricité (2019) 37–44.

[30] M.M.H. Shandhi, et al., Estimation of instantaneous oxygen uptake during exercise and daily activities using a wearable cardio-electromechanical and environmental sensor, IEEE J. Biomed. Health Inform. 25 (2021) 634–646.

[31] M. Schmid, D. Rath, U. Diebold, Why and how savitzky-golay filters should be replaced, ACS Meas. Sci. Au 2 (2022) 185–196.

[32] A.E. Cetin, O.N. Gerek, Y. Yardimci, Equiripple FIR filter design by the FFT algorithm, IEEE Signal Process. Mag. 14 (1997) 60–64.

[33] P.-O. Åstrand, I. Ryhming, A nomogram for calculation of aerobic capacity (physical fitness) from pulse rate during submaximal work, J. Appl. Physiol. 7 (1954) 218–221.

[34] S.J. Strath, A.M. Swartz, D.R. Bassett, W.L. O'Brien, G.A. King, B.E. Ainsworth, Evaluation of heart rate as a method for assessing moderate intensity physical activity, Med. Sci. Sport. Exerc. 32 (2000) S465–S470.

[35] L.R. Keytel, J.H. Goedecke, T.D. Noakes, H. Hiiloskorpi, R. Laukkanen, L. van der Merwe, E.V. Lambert, Prediction of energy expenditure from heart rate monitoring during submaximal exercise, J. Sports Sci. 23 (2005) 289–297.

[36] M.J. Sullivan, J.D. Knight, M.B. Higginbotham, F.R. Cobb, Relation between central and peripheral hemodynamics during exercise in patients with chronic heart failure. Muscle blood flow is reduced with maintenance of arterial perfusion pressure, Circulation 80 (1989) 769–781.

[37] F. Wilcoxon, Individual comparisons by ranking methods, Biom. Bull. 1 (1945) 80.
[38] T.K. Koo, M.Y. Li, A guideline of selecting and reporting intraclass correlation coefficients for reliability research, J. Chiropr. Med. 15 (2016) 155–163.
[39] D.M. Cooper, C. Berry, N. Lamarra, K. Wasserman, Kinetics of oxygen uptake and heart rate at onset of exercise in children, J. Appl. Physiol. 59 (1985) 211–217.
[40] H.A.M. Daanen, R.P. Lamberts, V.L. Kallen, A. Jin, N.L.U. Van Meeteren, A systematic review on heart-rate recovery to monitor changes in training status in athletes, Int. J. Sport. Physiol. Perform. 7 (2012) 251–260.
[41] J.L. Sabino-Carvalho, L.C. Vianna, Altered cardiorespiratory regulation during exercise in patients with Parkinson's disease: A challenging non-motor feature, SAGE Open Med. 8 (2020).
[42] S. Kitae, Y. Murata, N. Tachiki, M. Okazaki, T. Harada, S. Nakamura, Assessment of cardiovascular autonomic dysfunction in multiple system atrophy, Clin. Auton. Res. 11 (2001) 39–44.
[43] F. Braga, T. Bessa, M. Cunha, R.B. Paiva, R. Stein, Exercise-induced complete atrioventricular block resulting in cardiac arrest: a case report and review of diagnostic strategies, Eur. Hear. J. - Case Rep. 8 (2024).
[44] M.M. Abudiab, M.M. Redfield, V. Melenovsky, T.P. Olson, D.A. Kass, B.D. Johnson, B.A. Borlaug, Cardiac output response to exercise in relation to metabolic demand in heart failure with preserved ejection fraction, Eur. J. Hear. Fail. 15 (2013) 776–785.
[45] W. Kosowski, K. Aleksandrowicz, Hypertensive response to exercise as an early marker of disease development, Biomedicines 13 (2024) 30.
[46] S. Que, R. Poelarends, P. Veltink, M. Vollenbroek-Hutten, Y. Wang, Accelerometry-based energy expenditure estimation during activities of daily living: A comparison among different accelerometer compositions, 2025, arXiv preprint arXiv:2502.10112.
[47] S. Rang, et al., Modelflow: a new method for noninvasive assessment of cardiac output in pregnant women, Am. J. Obstet. Gynecol. 196 (2007) 235.e1–235.e8.
[48] C. Scott, Misconceptions about aerobic and anaerobic energy expenditure, J. Int. Soc. Sport. Nutr. 2 (2005).
[49] N. Kawel-Boehm, A. Maceira, E.R. Valsangiacomo-Buechel, et al., Normal values for cardiovascular magnetic resonance in adults and children, J. Cardiovasc. Magn. Reson. 17 (2015) 29.
[50] N. Naderi, Hemodynamic study, Pr. Cardiol. (2022) 201–216.
[51] T.A. Goers, The Washington Manual of Surgery, Lippincott Williams and Wilkins, 2010.
[52] J.E. Hall, Guyton and Hall Textbook of Medical Physiology, fourteen ed., Elsevier, 2020.